\documentclass[10pt]{article}

\usepackage[margin=1in]{geometry}
\usepackage[T1]{fontenc}
\usepackage[utf8]{inputenc}
\usepackage{lmodern}
\usepackage{microtype}
\usepackage{graphicx}
\usepackage{booktabs}
\usepackage{multirow}
\usepackage{amsmath,amssymb,bm}
\usepackage{siunitx}
\usepackage{subcaption}
\usepackage[numbers,sort&compress]{natbib}
\usepackage{xcolor}
\usepackage{enumitem}
\usepackage{url}
\usepackage{hyperref}
\usepackage{tikz}
\usetikzlibrary{arrows.meta,positioning,fit}

\graphicspath{{figures/}}
\hypersetup{
  colorlinks=true,
  linkcolor=blue!55!black,
  citecolor=green!45!black,
  urlcolor=blue!65!black,
  pdftitle={Toward Goal-Agnostic Joint-Embedding Predictive Control of Partial Differential Equations}
}
\setlist{nosep,leftmargin=*}

\title{
  Toward Goal-Agnostic Joint-Embedding\\
  Predictive Control of Partial Differential Equations
}

\author{
  \textbf{Jonathan Gallagher}\textsuperscript{\textdagger}\\
  University of Waterloo\\
  \texttt{jonathan.gallagher@uwaterloo.ca}
  \and
  \textbf{Roberto Guglielmi}\\
  University of Waterloo\\
  \texttt{roberto.guglielmi@uwaterloo.ca}\\[0.75em]
  \small\textsuperscript{\textdagger}Corresponding author
}
\date{\today}

\begin{document}
\maketitle

\begin{abstract}
We present a goal-agnostic control framework for partial differential equations
(PDEs) built around an end-to-end joint-embedding predictive architecture (JEPA). A lightweight 2D vision-transformer (ViT) and action-conditioned latent dynamics are trained offline without a reward or downstream goal, before being frozen and reused by a model-predictive path integral (MPPI) controller. We minimize a control objective in the latent space, initially expressed via the $L^2$ distance and additionally illustrate the benefit of recasting the control objective in terms of an explicit physical observable when available. By instead minimizing the tracking error for a learned linear kinetic-energy (KE) probe on the frozen latent-state rollouts, we demonstrate the ability to reproduce the control of held-out trajectories with $R^2=0.989$, while requiring no change to the underlying world model. For a controlled 2D Navier--Stokes benchmark, using a KE-probe within MPPI planning improves the mean native reward from $-12.08\pm0.86$ for latent-$L^2$ tracking to $-10.90\pm0.91$ (95\% CI), all while lowering last-quarter velocity-field
RMSE from $0.0765$ to $0.0692$.  Across three intentionally withheld, dissimilar, aperiodic targets, KE planning lowers late field RMSE by $53\%$ relative to latent-$L^2$ planning ($0.0220$ versus $0.0469$), winning across 30 paired comparisons. The same frozen model also supports stabilization around a steady-state configuration via direct regulation of KE, achieving $2.7\%$ mean relative error. While the latent probe proves brittle to measurement noise and missing pixels, our findings support the claim that latent dynamics can remain flexible and goal-agnostic, particularly when calibrated observables (granted they guarantee unique continuation) are a suitable objective for state control.
\end{abstract}

\section{Introduction}

Partial differential equations (PDEs) are crucial modeling tools, arising in many applications; ranging from climate modeling, biological processes, fluid--structure interactions, chemical process monitoring, and more. Developing control strategies for PDE-systems can be challenging, the state-space is often high dimensional, the relationships between quantities often nonlinear, and computational costs which scale super-linearly with respect to the number of state variables. As such, the evaluation of candidate control-sequences often requires computationally costly numerical simulation, typically precluding online planning and control. Extant methods for PDE control, such as model-based control, learned surrogate control, or reinforcement learning approaches, have been applied with varying degrees of success. Classical model-based controllers require problem-specific analysis, or otherwise clever construction of reduced-order models (ROMs) \cite{brunton2015closedloop}. Learned surrogate approaches introduce neural networks as approximations of the evolution operator for system dynamics \cite{holl2020pdecontrol,hwang2022operatorcontrol,mowlavi2023pinncontrol}, but quite often the surrogate is trained on the task of state-space forward-simulation and optimized against a specific objective, inheriting the challenges of full state reconstruction and requiring retraining when objectives change \cite{wei2024diffphycon,wu2022lepde}. Reinforcement learning has demonstrated excellent performance on well-defined tasks \cite{bucci2019chaotic, farahmand2017pdecontrol, rabault2019flowcontrol,verma2018collective}, but suffers from the need to carefully select the reward function, limiting the capacity for generalization.

Joint-embedding predictive architectures (JEPAs) instead focus on learning how applied actions map between elements of a learned latent space or ``World Model''. This is accomplished without necessitating state reconstruction at any point \cite{assran2023ijepa}. Recent work in latent world modeling has leveraged reconstruction-free features for
zero-shot planning primarily in visual domains \cite{zhou2024dinowm}, while existing latent PDE
work indicates that compressed dynamics may accelerate both simulation and inverse
optimization \cite{wu2022lepde}. We believe JEPA-style latent world models offer unique benefits for PDE problems. In particular, training with random masking can potentially enable control subject to both partially observed, or corrupt measurements. Perhaps of the greatest interest, performing planning and control entirely in a learned latent representation allows for goal-independent model-predictive control without the computational burden of state reconstruction. The learned dynamics model remains agnostic to the run-time goal, since changing this objective at the time of deployment does not require retraining either the encoder or predictor.

Latent planning is attractive for many reasons: its computational efficiency, and goal agnosticism being the two most obvious. Unfortunately, a latent representation structured such that it prioritizes predictability and action sensitivity is not automatically a calibrated metric space, and $L^2$ distance between embeddings can distort both the scale and ordering of physical field error. We employ a temporal-straightening penalty in an attempt to improve the latent rollout geometry \cite{wang2026temporalstraighteninglatentplanning}, but our experiments show that regularization alone does not suffice to solve this issue entirely.
Instead, we fit a lightweight latent-probe corresponding to a physical state observable and form
the tracking error \emph{after} that readout, in physical units. Our observable is chosen carefully to guarantee unique continuation, and we observe that implementing controls through this observable probe substantially improves closed-loop control, all
without retraining or sacrificing the flexibility of the world model formulation.

Recently, JEPA world models have been successfully applied for planning in visual manipulation,
navigation, locomotion, and aerial domains \cite{assran2025vjepa2,zhou2024dinowm,maes2026leworldmodel,
zhang2026deltajepa, rao2026skyjepalearninglonghorizonworld}. To our knowledge, a JEPA-like joint-embedding world model has yet to be applied to the closed-loop control of PDE systems. This work aims to demonstrate the potential for joint-embedding architectures in PDE control, showing that a goal-independent, joint-embedding world model coupled with a downstream action-conditioned predictor can be a uniquely dynamic controller, capable of natively supporting multiple PDE-control objectives.

We validate this framework in the PDE Control Gym two-dimensional
Navier--Stokes environment, comparing against existing Reinforcement-Learning (RL) approaches for lid-driven cavity flow control \cite{bhan2024pdecontrolgym}. We train an end-to-end joint-embedding architecture consisting of a ViT, action-conditioned predictor, and lightweight physical probe. To prevent collapse and encourage a well-structured latent space, we employ a VICReg penalty
\cite{bardes2022vicreg}, temporal straightening \cite{wang2026temporalstraighteninglatentplanning} and Delta-JEPA-like latent action decoding \cite{zhang2026deltajepa} during training. We employ a sampling-based MPPI controller with the aforementioned frozen backbone to facilitate our latent planning and control \cite{williams2017mppi}. 

\vspace{0.5cm}
\textbf{Our Contributions: }

\begin{enumerate}
  \item To our knowledge, we introduce the first implementation of closed-loop PDE-control using a joint-embedding latent world model, capable of learning action-conditioned dynamics independently of the downstream control task.
  \item We introduce an observable-aligned controller: a cheap-to-train
  linear probe estimates kinetic energy from predicted latents, and MPPI evaluates tracking
  error in that physical scalar rather than using raw latent distance.
  \item On the PDE Control Gym benchmark, this improves both
  native reward and velocity-field RMSE over the controller using latent-$L^2$
  cost.  Across three distinct aperiodic signals, it reduces late field RMSE
  by $53\%$, with improvement in all 30 paired episodes.
\end{enumerate}

\section{Related Work}

Planning in a learned latent space predates the introduction of joint-embedding architectures. For example, Embed-to-Control learns
locally linear latent dynamics for control from raw images \cite{watter2015e2c}; World Models \cite{ha2018worldmodels} and PlaNet \cite{hafner2019planet} either plan or learn behaviors inside learned latent dynamics, with Dreamer extending to latent imagination \cite{hafner2020dreamer} and PETS establishing sampling-based MPC over learned probabilistic dynamics \cite{chua2018pets}.
Closest to our work are TD-MPC and TD-MPC2, which run MPPI over a
reconstruction-free latent dynamics model trained with a latent consistency
loss \cite{hansen2022tdmpc, hansen2024tdmpc2}, however in both cases latent spaces are shaped by task rewards and value learning, whereas a joint-embedding approach like ours is trained explicitly with no task signal at all. MuZero also plans in a value-equivalent latent space
without reconstruction \cite{schrittwieser2020muzero}.  Our contribution
transfers this latent-planning recipe to PDE dynamics while keeping the
representation goal-agnostic.

JEPA models seek to learn an action-conditioned world model, which describes how varying control actions transition between latent embeddings of both the current and target state, rather than working in the input space \cite{lecun2022path, assran2023ijepa}. Generally, this is accomplished by self-supervised training, combining masked prediction with vision transformers \cite{dosovitskiy2021vit, he2022mae} and a lagged, target encoder using an exponential moving average (EMA) as an anti-collapse mechanism \cite{grill2020byol}. For end-to-end models such as ours, additional anti-collapse measures must be taken, VICReg supplies explicit variance and covariance regularization against collapse \cite{bardes2022vicreg}.  DINO-WM demonstrates that pretrained non-generative visual features can support zero-shot planning
\cite{zhou2024dinowm}; V-JEPA~2 post-trains an action-conditioned predictor on
robot interaction data and plans manipulation from image goals
\cite{assran2025vjepa2}; LeWorldModel learns an end-to-end JEPA for manipulation, navigation, and locomotion tasks \cite{maes2026leworldmodel}; SkyJEPA extends JEPA world models to long-horizon quadrotor control
\cite{rao2026skyjepalearninglonghorizonworld}. Delta-JEPA targets the complementary control problem of action-insensitive latent dynamics by decoding actions from latent differences \cite{zhang2026deltajepa}. While existing work establishes JEPA world modeling as a promising avenue for visual and robotic domains, to our knowledge, ours is the first to apply a JEPA world
model to closed-loop PDE control explicitly.

Learned PDE surrogates commonly evolve either full fields or compressed states.
Differentiable-solver control \cite{holl2020pdecontrol}, operator-learning
control \cite{hwang2022operatorcontrol}, PINN-based optimal control
\cite{mowlavi2023pinncontrol}, and diffusion-based control
\cite{wei2024diffphycon} optimize actions through a surrogate coupled to a
full state-space objective, while neural operators have shown promise in learning the gain computations
of PDE backstepping directly \cite{bhan2024neuralopbackstepping}. LE-PDE proposes a latent-evolution approach that targets both accelerated simulation and inverse optimization \cite{wu2022lepde}.  Closest in spirit to our setting are Koopman and reduced-order methods performing control in learned low-dimensional spaces. Examples of this include deep Koopman embeddings \cite{lusch2018koopman},
Koopman-based MPC \cite{korda2018koopmanmpc, peitz2019koopmanpde}, SINDy-MPC
\cite{kaiser2018sindympc}, latent MPC of unsteady fluid flows
\cite{morton2018deepfluid}, and deep-ROM latent feedback control
\cite{tomasetto2024latentfeedback}. Our focus differs from existing latent evolution works by adopting the JEPA / World Modeling framework, removing the field decoder entirely and fully separating the learned dynamics from
the deployment goal.

\section{Learned Control Formulation}

\subsection{Learning action sensitive dynamics}

Let a controlled field $U_t$ evolve as $U_{t+1}=F(U_t,a_t)$ where $F$ is a transition function defined in state-space, conditioned both upon the current state and the applied control $a_t$.  We learn a patched ViT encoder
$E_\theta$, and action-conditioned latent predictor $g_\phi$ consisting of a lightweight, history conditioned multi-layer perceptron (MLP). Together, the encoder $E_\theta$ and the predictor $g_\phi$ act as a latent surrogate for the state transition function $F$, allowing for latent-state transitions to be understood in terms of the actions which caused them. The architecture is trained end-to-end using only offline state transitions and their corresponding control actions. The full training procedure is depicted in Figure~\ref{fig:training_process}. 

Once trained, the weights of both the encoder $E_\theta$ and action-conditioned predictor $g_\phi$ remain frozen. To perform control, a sampling based MPPI planner then supplies a total of 512 candidate action sequences, whose corresponding latent state transitions are probed through auto-regressively rolling out the learned predictor $g_\phi$.  This formulation allows for a goal to be specified in terms of target states or derived quantities entirely within the MPPI planner at the time of deployment. By leveraging the learned relationship between proposed actions and latent space dynamics, the world model presents a cheap to inference, flexible surrogate for the complete PDE system.

\subsection{Control objectives}

Our primary emphasis is the flexibility of the planning machinery, which is unique in that it supports arbitrary control objectives. To demonstrate this, we study three distinct control scenarios:
\begin{enumerate}
 \item \textbf{Time-dependent tracking with action penalty.} A state-space target sequence over a short time horizon $H$ is provided as $q^\star_{t+1:t+H}$, and subsequently encoded as the latent state trajectory $z^\star_{t+1:t+H}$. The controller is tasked with learning to track the target latent sequence while penalizing wasteful actions. On this task, we demonstrate comparable performance to existing RL baselines despite not being trained in a cost-aware manner.
  \item \textbf{Time-dependent tracking.}  A target state-space sequence over a short time horizon $H$ is provided as $q^\star_{t+1:t+H}$, and encoded as latent states $z^\star_{t+1:t+H}$. The controller must find controls which track the target sequence as closely as possible.
  \item \textbf{Fixed-target stabilization.} A specified state configuration $q^\star$ is supplied and encoded only once. The corresponding encoded latent $z^\star$ is treated as a constant target, resulting in an equilibrium objective. The controller must find controls which stabilize around the target latent embedding.
\end{enumerate}

Only the user-specified cost changes between these tasks, meaning the action-conditioned
latent dynamics model and MPPI optimizer are otherwise unchanged.

\subsection{PDE Control Gym benchmark}

We validate the model on the control of the 2D Navier--Stokes equations. Using the PDE Control Gym two-dimensional
Navier--Stokes environment, we study using boundary control for a lid-driven cavity flow example
\cite{bhan2024pdecontrolgym}. 

\begin{subequations}
\begin{align}
\nabla\cdot \mathbf{u} &= 0, \\
\frac{\partial \mathbf{u}}{\partial t}
+ \mathbf{u}\cdot\nabla\mathbf{u}
&=
-\frac{1}{\rho}\nabla p
+\nu\nabla^2\mathbf{u}.
\end{align}
\end{subequations}
Using the published configuration from PDE Control Gym, the domain is a unit square, whereupon we denote the spatial variable $\mathbf{x} = (x,y) \in \mathcal{X} = [0,1]\times[0,1]$. Our 2D velocity field is represented as $\mathbf{u} = (u,v) : \mathcal{X}\times [0,T] \to \mathbb{R}^2$, $\nu$  the kinematic viscosity, $\rho$  the fluid density, and $p$ the pressure field. For the control task, we study the effects of a boundary control applied along the top boundary  as $\mathbf{u}(x,1,t) = a(x,t),~ \forall x \in [0,1]$. Remaining boundary conditions are Dirichlet ``no-slip'' boundaries where velocity is set to zero.  

For the purposes of numerical simulation, we discretize our domain with $\Delta x=\Delta y=0.05$ into a $21\times21$ grid. The simulator employs a timestep $\Delta t=10^{-3}$, a total time horizon of $T=0.2$, a kinematic viscosity $\nu=0.1$, fluid density $1$, and 2,000 pressure iterations.  The observation
is the velocity field
$U_t=(u_t,v_t)\in\mathbb{R}^{21\times21\times2}$, where $(u_t,v_t)$ represent the $x$ and $y$ components of velocity respectively.  A scalar action
$a_t\in[0,4]$ sets a controllable Dirichlet condition (in the positive $x$ direction) along the upper boundary. For benchmarking purposes we seek to recover a generating control action which is unknown to the learned controller at the time of inference. That is to say the controller must learn the generating control from embedded state observations only. 

The default reward function consists of time-dependent tracking with action-penalty. As written, it asks that the controller balance the cost of tracking without using needlessly large controls, taking the form: 
\begin{equation}
 r_t = -\frac{1}{2N_xN_y}\lVert U_t-U_t^\star\rVert_2^2
       -\frac{\lambda_a}{2}(a_t-a_t^{\mathrm{ref}})^2,
 \label{eq:reward}
\end{equation}
where $N_x=N_y=21$ denotes the spatial resolution, $\lambda_a=0.1$ the scalar weight applied to the action penalty, and $a_t^{\mathrm{ref}}=2$ the reference action, from which deviations are penalized. In subsequent tables, we report both the cumulative reward and state-space RMSE. To make learning state transitions more tractable, our controller acts once per two native solver steps. That is to say we employ 99 control actions, holding each for twice the benchmark's native timestep.

\section{Goal-Agnostic JEPA Control Implementation}

\subsection{Masked observations}

\begin{figure}[htbp]
    \centering
    \includegraphics[width=1\linewidth]{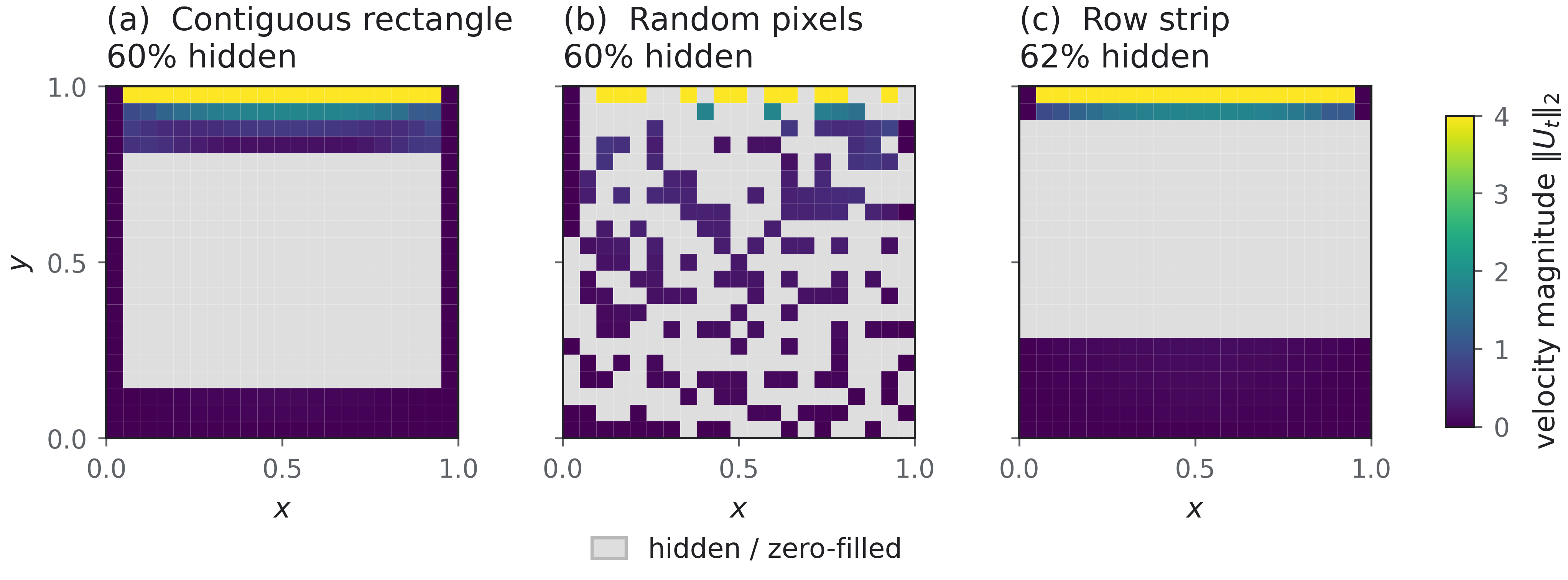}
    \caption{Masking protocol: contiguous rectangles, random pixels, and row
    strips, each shown hiding roughly 60\% of the field. Grey denotes masked pixels.}
    \label{fig:masking_protocol}
\end{figure}

When training our encoder (lightweight ViT), state observations are randomly masked. For a given state there is probability $0.15$ that it
is fully observed during training; otherwise the masking fraction is sampled uniformly on the range $[0.3,0.9]$. The type of mask is selected from one of either contiguous rectangles, random pixels, or row strips with respective probabilities $0.5$, $0.3$, and $0.2$. The purpose of training using masked observations is not only to ensure robustness to partial observation, but to emphasize the most important semantic content in a given snapshot. During evaluation, the observed state is either fully observed, corrupted by zero-mean Gaussian noise, or missing $5$--$25\%$ of its pixels at random. The target state on the other hand is always fully observed. Through this convention we interpret the target state as a specified objective rather than a measured quantity. An example of the masking protocol is shown in Figure~\ref{fig:masking_protocol}.

\subsection{Learning latent dynamics}

Let $E_\theta$ denote a masked field encoder and let $z_t=E_\theta(U_t,m_t)$ denote the latent encoding of state $U_t$ given masking tokens $m_t$. The encoder consists of a two-dimensional ViT, which produces a 256-dimensional latent embedding
output. The encoder employs $3\times3$ patches, four transformer blocks, and four attention heads. The predictor $g_\phi$ takes a history of four latent states and their actions as input before auto-regressively rolling out $H_{\mathrm{train}} = 12$ latent steps. The predictor consists of a two-layer MLP with 512 hidden units per layer, trained to predict latent residuals, or the difference between subsequent latent states rather than a sequence of full states, such that:
\begin{equation}
    g_\phi(a_{t-K:t},\hat z_{t-K:t}) = \Delta \hat{z}_t = \hat z_{t+1} - \hat z_t,
\end{equation}
from which $z_{t+1}$ is computed as
\begin{equation}
    z_{t+1} = z_t + g_\phi(a_{t-K:t},\hat z_{t-K:t}),
\end{equation}
and the subsequent latent states can be reconstructed straightforwardly via autoregressive rollout.

\begin{figure}[htbp]
    \centering
    \includegraphics[width=0.92\linewidth]{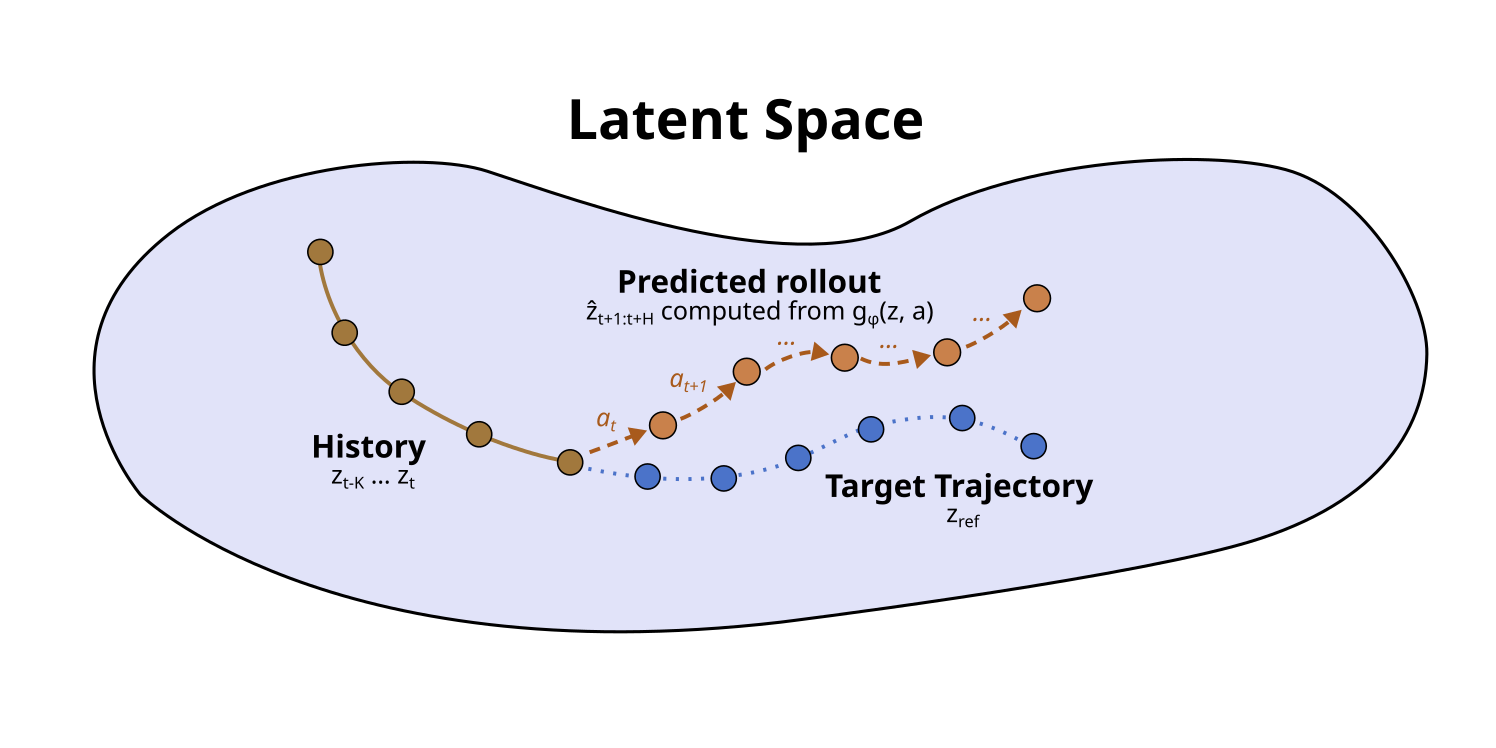}
    \caption{Overview of latent-space planning. The encoder maps observations, and target states into a latent representation. The learned predictor generates a future rollout conditioned on candidate actions. The distance between the predicted rollout and the target trajectory is minimized with respect to a chosen measure.}
    \label{fig:predictor}
\end{figure}

For any given training sample, we first encode the masked history, after which we encode a second, unmasked sequence of target states, represented by:
\begin{equation}   
 z^{\mathrm{context}}_{t-K+1:t}=E_\theta(U_{t-K+1:t},m),\qquad
 z^{\mathrm{target}}_{t:t+H_{\mathrm{train}}}=E_\theta(U_{t:t+H_{\mathrm{train}}},\mathbf{1}).
\end{equation}
Where context denotes the previous state history embedding with a window of size $K$, and where $H_{\mathrm{train}}$ is the prediction horizon, with $\mathbf{1}$ a key denoting the absence of masking tokens for target states. Once both history and target states have been embedded, the predictor $g_\phi$ rolling forward from $z^{\mathrm{context}}$, is compared against the target states (shown in Figure~\ref{fig:predictor}), with the rollout loss $\mathcal{L}_{\mathrm{roll}}$ computed as:
\begin{equation}
\displaystyle
 \mathcal{L}_{\mathrm{roll}}
 = \frac{1}{H_\mathrm{train}}\sum_{h=1}^{H_\mathrm{train}}
 \lVert \hat z_{t+h}-z^{\mathrm{target}}_{t+h}\rVert_2^2.
\end{equation}

Gradients from $\mathcal{L}_{\mathrm{roll}}$ are permitted to flow through each of the predictor, the masked-history encoder branch, and the unmasked-future encoder branch, to encourage simultaneous learning of both a better latent representation and predictor.

To discourage representational collapse, we employ both VICReg variance $(\mathcal{L}_{\mathrm{var}})$ and
covariance $(\mathcal{L}_{\mathrm{cov}})$ penalties \cite{bardes2022vicreg}. To do so, we let
$Z\in\mathbb{R}^{N\times D}$ denote a collection of $N$ latent vectors
of dimension $D$, where $\widetilde{Z}=Z-\mathbf{1}\bar{z}^{\top}$ denotes the same samples batch-centered. Define the empirical covariance matrix
as
\begin{equation}
    C(Z)=\frac{1}{N-1}\widetilde{Z}^{\top}\widetilde{Z}.
\end{equation}
Once having computed the covariance matrix as above, the VICReg penalties are applied as
\begin{align}
    \mathcal{L}_{\mathrm{var}}(Z)
    &=
    \frac{1}{D}\sum_{j=1}^{D}
    \max\left\{
        0,\,
        \gamma-\sqrt{C(Z)_{jj}+\varepsilon}
    \right\},
    \\
    \mathcal{L}_{\mathrm{cov}}(Z)
    &=
    \frac{1}{D}
    \sum_{\substack{i,j=1\\i\neq j}}^{D}
    C(Z)_{ij}^{2},
\end{align}
where $\gamma$ is the target standard deviation and $\varepsilon>0$ a small scalar value
included for numerical stability. We apply these penalties symmetrically
to the predicted and target representations such that:
\begin{align}
    \mathcal{L}_{\mathrm{var}}
    &=
    \mathcal{L}_{\mathrm{var}}(Z^{\mathrm{pred}})
    +
    \mathcal{L}_{\mathrm{var}}(Z^{\mathrm{real}}),
    \\
    \mathcal{L}_{\mathrm{cov}}
    &=
    \mathcal{L}_{\mathrm{cov}}(Z^{\mathrm{pred}})
    +
    \mathcal{L}_{\mathrm{cov}}(Z^{\mathrm{real}}).
\end{align}
By rewarding variance, the VICReg penalty ensures that variation is significant along each latent
coordinate, while the corresponding covariance penalty discourages representational redundancy. 

To ensure the latent space remains conducive to action-conditioned planning, an auxiliary decoder $d_\psi$ consisting of a lightweight MLP is tasked with reconstructing the applied
action from adjacent displacements in the latent-encoded target sequence, inspired by Delta-JEPA \cite{zhang2026deltajepa}. The resulting penalty
$\mathcal{L}_{\Delta a}$ encourages the latent space to remain organized
in a manner suitable for recovering the applied actions:
\begin{equation}
    \mathcal{L}_{\Delta a}
    =
    \frac{1}{H_{\mathrm{train}}}\sum_{h=1}^{H_{\mathrm{train}}}
    \left\|
        d_\psi\left(
            z^{\mathrm{real}}_{t+h}
            -
            z^{\mathrm{real}}_{t+h-1}
        \right)
        -
        a_{t+h-1}
    \right\|_2^2.
\end{equation}
Lastly, we employ a temporal-straightening penalty ($\mathcal{L}_{\mathrm{str}}$) as in \cite{wang2026temporalstraighteninglatentplanning}. This penalty is evaluated only on the unmasked, encoded ground-truth sequence, effectively discouraging successive latent displacements from changing direction, instead favoring locally straight temporal trajectories:

\begin{equation}
\mathcal{L}_{\mathrm{str}}
=
\frac{1}{B(M-2)}
\sum_{b=1}^{B}\sum_{t=1}^{M-2}
\left[
1-\cos_{\varepsilon}
\left(
\Delta z_{b,t},
\Delta z_{b,t+1}
\right)
\right],
\qquad
\Delta z_{b,t}
=
z_{b,t+1}-z_{b,t}.
\end{equation}
Here, \(B\) is the batch size, \(M\) is the number of latent
states in each rollout, and the numerically stabilized cosine similarity is
\begin{equation}
\cos_{\varepsilon}(a,b)
=
\frac{a^{\top}b}
{\max(\lVert a\rVert_2,\varepsilon)\,
 \max(\lVert b\rVert_2,\varepsilon)}.
\end{equation}

\begin{figure}[htbp]
    \centering
    \includegraphics[width=0.9\linewidth]{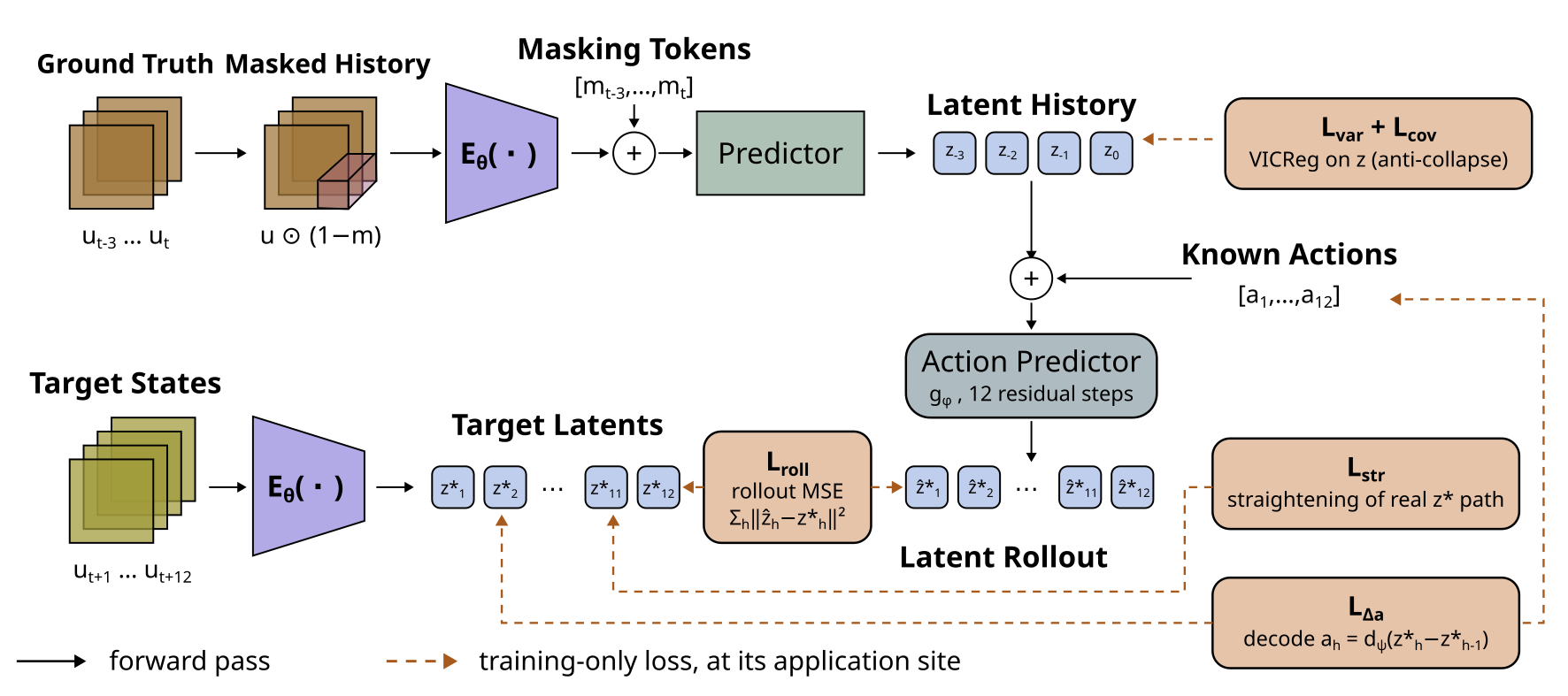}
    \caption{ Our joint-embedding training process, in which ground-truth and partially masked
    observations are encoded and rolled forward before their latents are compared.}
    \label{fig:training_process}
\end{figure}

Combining each of these terms, the total composite loss reads:

\begin{equation}
\mathcal{L}_{\mathrm{total}}
= \mathcal{L}_{\mathrm{roll}}
+ \lambda_{\mathrm{var}}\,\mathcal{L}_{\mathrm{var}}
+ \lambda_{\mathrm{cov}}\,\mathcal{L}_{\mathrm{cov}}
+ \lambda_{\mathrm{str}}\,\mathcal{L}_{\mathrm{str}}
+ \lambda_{\Delta a}\,\mathcal{L}_{\Delta a}.
\end{equation}

with weights $\lambda_{\mathrm{var}}=1.0$,
$\lambda_{\mathrm{cov}}=1.0$, $\lambda_{\mathrm{str}}=0.1$, and
$\lambda_{\Delta a}=10$. The full training procedure with loss terms is shown in Figure~\ref{fig:training_process}.

\newpage 
\subsection{Latent MPPI}\label{sec:mppi}

Given a candidate action sequence $A=(a_0,\ldots,a_{H-1})$, the frozen
predictor $g_\phi$ is used to generate the corresponding action-conditioned latent
rollout $\hat z_{t+1:t+H}$.  At runtime, our MPPI sampling based controller maintains a nominal plan
$\bar A^{(r)}$, where $r$ indexes the refinement iteration. At each refinement iteration, we sample $N_s$ perturbed candidates around the nominal action:
\begin{equation}
 \epsilon_i^{(r)}
 \sim\mathcal{N}(0,\sigma^2 I_H),
 \qquad
 A_i^{(r)}
 =
 \Pi_{[a_{\min},a_{\max}]^H}
 \left(\bar A^{(r)}+\epsilon_i^{(r)}\right),
 \qquad i=1,\ldots,N_s,
 \label{eq:mppi-sampling}
\end{equation}
where $\Pi$ denotes elementwise clipping to the admissible action interval.
Each candidate is propagated through $g_\phi$ at which point the goal is specified by way of an assigned cost $J_i^{(r)}=J(A_i^{(r)})$. Once a best control is selected for a given iteration, it becomes the nominal control for the next, iteratively refining the proposed action. For our case, we use 3 iterative refinement steps. For example, the conventional latent-space tracking with an action penalty scores action candidates based on the following: 
\begin{equation}
 J_{\mathrm{lat}}(A)
 =
 \alpha\sum_{h\in\mathcal{H}}
 \lVert \hat z_{t+h}-z^\star_{t+h}\rVert_2^{2}
 +
 \lambda_a\sum_{h=0}^{H-1}\left(a_h-a^{\mathrm{ref}}\right)^2,
 \label{eq:latent-cost}
\end{equation}
where $\mathcal{H}\subseteq\{1,\ldots,H\}$ selects a subset of rollout states upon which the score is computed.  By restricting $\mathcal{H}$ to later states, one can weight tracking toward a later interval, which is particularly useful for our setting, where boundary actuation requires time to propagate into the
domain.

In this work, we explore an alternative interface for latent tracking, through which we replace the latent residual in Equation~\eqref{eq:latent-cost} with an error term computed from a
calibrated scalar observable. Let $q_\omega$ denote a learned readout which maps
a predicted latent $\hat{z}_{t+h}$ to a particular physical quantity $q^\star_{t+h}$ the
corresponding reference value. In a familiar manner, the planner evaluates candidate actions using
\begin{equation}
 J_q(A)
 =
 \beta\sum_{h\in\mathcal{H}}
 \left[
 q_\omega(\hat z_{t+h})-q^\star_{t+h}
 \right]^2
 +
 \lambda_a\sum_{h=0}^{H-1}\left(a_h-a^{\mathrm{ref}}\right)^2.
 \label{eq:observable-cost}
\end{equation}
For a fixed target this formulation reduces to regulation around a single scalar observable
value $q^\star$.  Section~\ref{sec:probe} gives the instantiation of $q_\omega$
used throughout our experiments. Otherwise, nothing in the planner depends on that
choice.

The distinction between the two costs comes down to how easily they are optimized over. As written, Equation~\eqref{eq:latent-cost} implicitly assumes
that Euclidean displacement in the learned representation ranks candidate
actions in a manner which is consistent with physical field error, and while joint-embedding
training encourages predictable, action-conditioned features, it does not enforce isometry in a strict manner. By design, Equation~\eqref{eq:observable-cost} demands less of the learned representation: the readout $q_\omega$ need only recover a single physical
observable, after which candidate rollouts are compared downstream in
physically calibrated units.

For either penalty $J=J_{\mathrm{lat}}$ or $J=J_q$, the candidates are assigned
normalized exponential weights
\begin{equation}
 w_i^{(r)}
 =
 \frac{
 \exp\!\left[
 -\bigl(J_i^{(r)}-\rho^{(r)}\bigr)/\lambda_T
 \right]
 }{
 \displaystyle\sum_{j=1}^{N_s}
 \exp\!\left[
 -\bigl(J_j^{(r)}-\rho^{(r)}\bigr)/\lambda_T
 \right]
 },
 \qquad
 \rho^{(r)}=\min_j J_j^{(r)},
 \label{eq:mppi-weights}
\end{equation}
where $\lambda_T>0$ is the MPPI temperature, a hyperparameter which determines how strongly weighted the best candidate is.  Subtracting $\rho^{(r)}$ improves
numerical stability without changing the normalized weights.  The nominal plan
is then updated to the weighted mean of the clipped candidates,
\begin{equation}
 \bar A^{(r+1)}
 =
 \sum_{i=1}^{N_s}w_i^{(r)}A_i^{(r)},
 \qquad
 \sum_{i=1}^{N_s}w_i^{(r)}=1,
 \label{eq:mppi-update}
\end{equation}
so that MPPI never directly executes the minimum-cost sampled candidate; the
argmin candidate is retained only for diagnostic purposes.

After the final refinement iteration $R$, only the first action
$\bar a_0^{(R)}$ is executed.  The remaining plan is shifted forward,
\begin{equation}
 \bar A_{\mathrm{next}}^{(0)}
 =
 \left(
 \bar a_1^{(R)},\ldots,
 \bar a_{H-1}^{(R)},
 \bar a_{H-1}^{(R)}
 \right),
 \label{eq:mppi-shift}
\end{equation}
to warm-start optimization at the next control step.

All tasks use $H=5$ latent steps, scoring all five rollout states.  We use $N_s=512$ candidates, $R=3$
refinement iterations, $\sigma=0.6$, $[a_{\min},a_{\max}]=[0,4]$, and
initialize the nominal plan with the reference action $a^{\mathrm{ref}}=2$.
Unless otherwise stated, the MPPI temperature is $\lambda_T=0.1$.  The tracking-only
latent configuration uses $(\alpha,\lambda_a)=(1,0)$.  For the native-reward
comparison we use $(\alpha,\lambda_a)=(0.05,0.05)$ for latent-$L^2$ tracking
and $(\beta,\lambda_a)=(1,0.05)$ for kinetic-energy tracking, with $\alpha$
tuned so that the latent tracking term is comparable in scale to the physical
reference penalty.  For the three-signal aperiodic suite we use
$(\beta,\lambda_a)=(1,0)$ and $\lambda_T=0.004$; since no other cost term
is active, this yields softmax weights identical to $(\beta,\lambda_T)=(25,0.1)$.  The
latent-$L^2$ baseline retains $\lambda_T=0.1$.

\subsection{Observable readout: a kinetic-energy probe}\label{sec:probe}

For the Navier--Stokes benchmark we instantiate $q_\omega$ with the spatially
averaged kinetic energy per unit mass of a $21\times21$ velocity field
$U=(u,v)$,
\begin{equation}
k(U)
=
\frac{1}{2(21)^2}
\sum_{i=1}^{21}\sum_{j=1}^{21}
\left(u_{ij}^{2}+v_{ij}^{2}\right),
\label{eq:ke}
\end{equation}
so that $q_\omega=\hat k_\omega$ and $q^\star_{t+h}=k(U^\star_{t+h})$ in
Equation~\eqref{eq:observable-cost}.  A frozen linear probe maps a
predictor-rollout latent $\hat z\in\mathbb{R}^{256}$ to that scalar,
\begin{equation}
\hat{k}_{\omega}(\hat z)
=
w^{\top}\hat z+b,
\label{eq:probe}
\end{equation}
with $(w,b)$ fitted by ridge regression on frozen predictor-rollout latents.
No JEPA parameters are updated and no field decoder is introduced; the probe
attains $R^2=0.9891$ on held-out trajectories
(Section~\ref{sec:ke}).

Kinetic energy is not a general substitute for field distance.  It suffices
here because a single, sign-constrained boundary actuator restricts reachable
states to a narrow response family on which the map from field to kinetic
energy is effectively injective, so that matching $k$ pins down the field. Throughout the manuscript, we retain velocity-field RMSE as the primary evaluation metric independent of the planner cost.

\section{Experimental Protocol}

\subsection{Offline data and optimization}

The offline training dataset contains 6,930 system trajectories, each consisting of 200 velocity fields and 199 corresponding actions. We use an $\approx 92/8$ train--validation split, with 6,306 trajectories assigned to training and 624 left for validation. Candidate actions are explicitly constrained to the range $[0,4]$. The training dataset combines a series of distinct action sequences, counterfactual branches where candidate signals are rolled out before splitting into several potential cases, piecewise-constant and bang-bang samples. The evaluation targets are the Control Gym benchmark reference and three custom aperiodic pulse sequences. The established aperiodic sample triplet is specifically audited for distance from the training dataset
(Section~\ref{sec:pulse}).

\begin{figure}[htbp]
    \centering
    \includegraphics[width=1\linewidth]{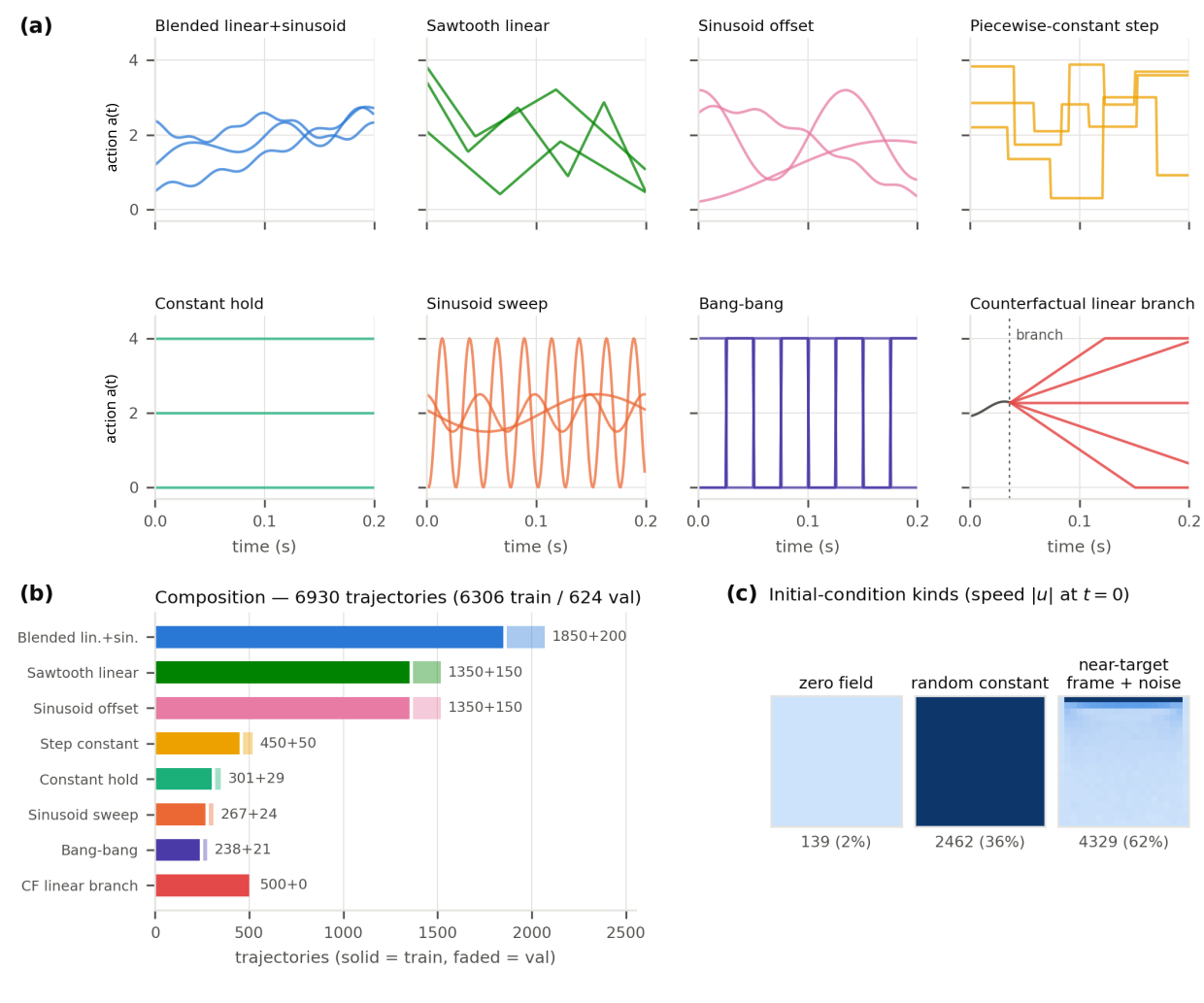}
    \caption{Composition of the offline training corpus. (a) Representative trajectories from the eight control-action families used for data generation. (b) Distribution of the 6,930 trajectories across action families and the training--validation split. Note: Counterfactual branches were not used in validation, because of complications in handling multiple potential branches. (c) Distribution of initial conditions among zero-field, random-constant, and noisy near-target states.}
    \label{fig:training_data}
\end{figure}
\begin{figure}[htbp]
  \centering
  \includegraphics[width=0.94\textwidth]{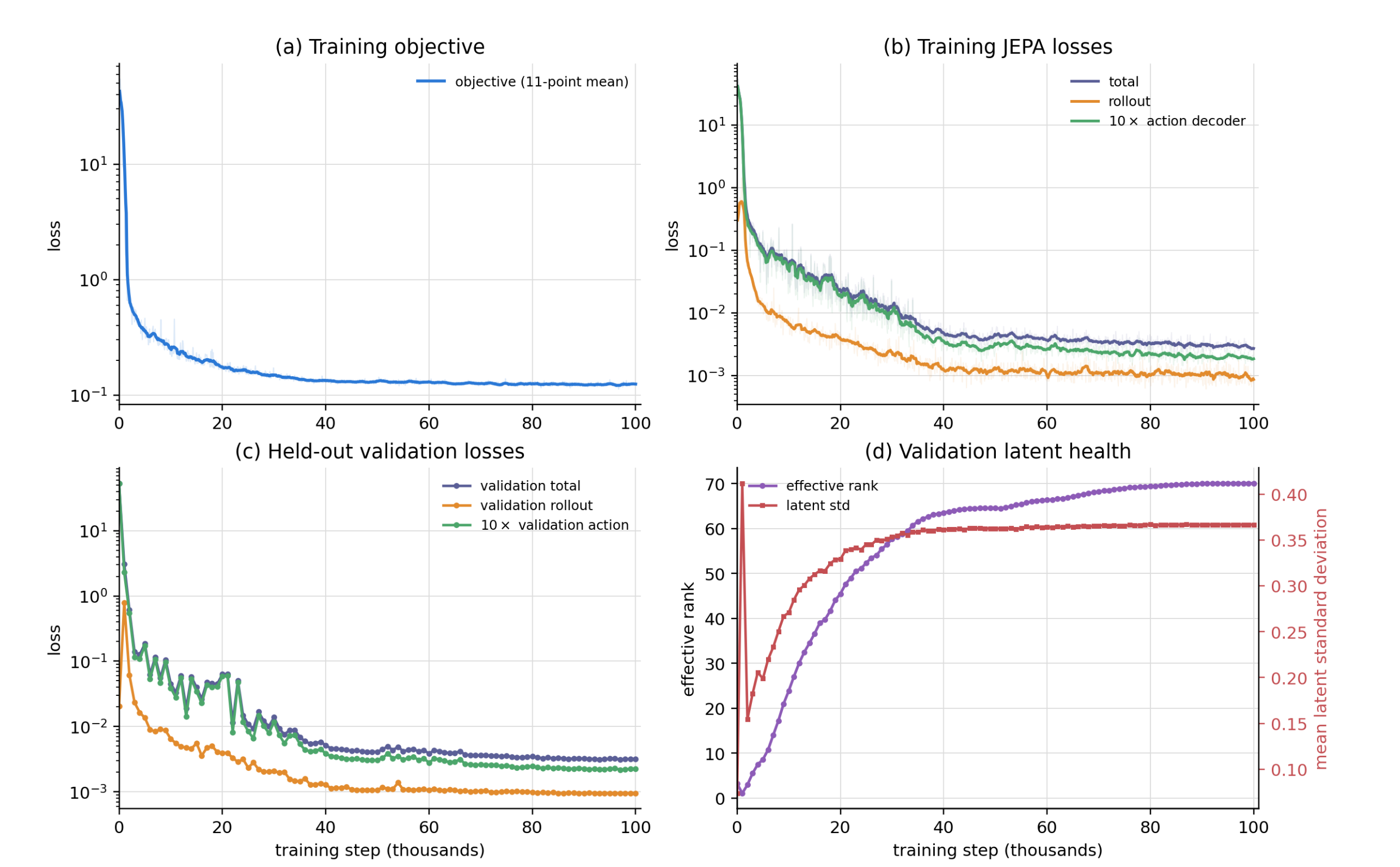}
    \caption{Training and validation diagnostics over the full 100k-step schedule. (a) The training objective learning quickly before plateau. (b) The JEPA rollout and action-decoding losses continue to improve throughout training. (c) Held-out losses can be observed to closely track the training losses. (d) The validation latent effective rank increases from approximately 1 to 70, while the mean coordinate-wise standard deviation undergoes a brief initialization transient before recovering and stabilizing. Together, the increasing rank and sustained latent variance indicate that the learned representation remains non-collapsed and progressively spreading information across more latent dimensions.}
  \label{fig:training}
\end{figure}

Our model is first trained for 50,000 optimizer steps with a batch size 128, learning rate $3\times10^{-4}$, weight decay $10^{-2}$, a 2,000-step warmup, cosine decay.  We then fine-tune for 50,000 more steps at a reduced peak learning rate of $6\times10^{-5}$ and a cosine decay with terminal learning rate zero. The resulting model is the single world model behind every result in this paper. Model training curves are shown in Figure~\ref{fig:training}. For all closed-loop control results, we present an average taken over ten spatially constant random initial velocity conditions sampled on $(-5,5)$. For the published-benchmark reward comparison, we extend this protocol to 50 random initial conditions to match the published count in Control Gym. All aggregate result figures depict 95\% CI unless stated otherwise.

\newpage
\section{Results}

We organize results by control objective at deployment. We first consider a direct
comparison of control discovery using raw latent-$L^2$ versus observable KE costs. Between examples, only the cost function and its scalar weight(s) are changed.  We report reconstructed
velocity-field RMSE independently of the planner cost so that it can be demonstrated that KE
tracking indeed establishes improved flow tracking in state space.

\subsection{Time-dependent target trajectories}

\subsubsection{PDE Control Gym benchmark}

\begin{figure}[htbp]
  \centering
  \includegraphics[width=0.98\textwidth]{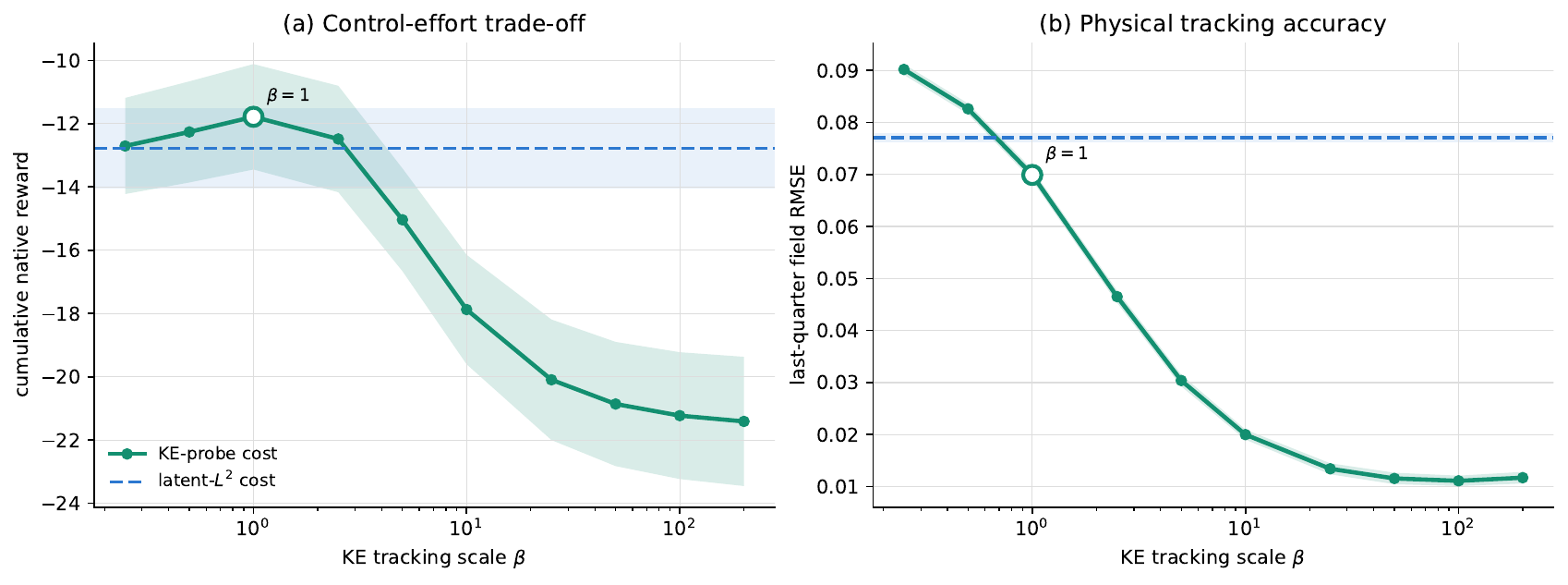}
  \caption{Clean linear-reference KE cost-scale screen over ten matched
  episodes (mean and 95\% CI).  The action-reference weight $\lambda_a$ remains $0.05$ throughout all trials.
  Dashed lines and blue bands show the matched latent-$L^2$ mean and 95\% CI.
  The circled $\beta=1$ operating point is used for the 50-episode benchmark
  comparison.}
  \label{fig:linear}
\end{figure}

We first sweep the KE scaling factor $\beta$ (in Equation~\eqref{eq:observable-cost}) with the scaling for the action-reference penalty
fixed at $\lambda_a=0.05$, consistent with the PDE Control Gym benchmark \cite{bhan2024pdecontrolgym}. We compare cumulative mean reward on ten matched clean episodes.  Figure~\ref{fig:linear}
shows a smooth control-effort frontier.  The latent-$L^2$ convention lies
inside this frontier at the reward-oriented choice $\beta=1$. We emphasize the evaluation of late-time tracking, using the mean RMSE computed over the final quarter of each $0.2\,\mathrm{s}$ episode, corresponding to the interval $t\in[0.15,0.20]\,\mathrm{s}$. Compared with latent-space planning, KE-based planning achieves both a higher mean reward ($-11.78$ versus $-12.78$) and a lower late-time field RMSE ($0.0699$ versus $0.0771$).  At $\beta=100$, it reaches late field RMSE $0.0111$, $46\%$ below even the latent tracking-only configuration's $0.0206$.

\begin{table}[htbp]
\centering
\caption{Clean linear-reference results.  Reward is higher when less negative;
parentheses give the planner tracking scale and action-reference weight.
Reward intervals are 95\% CIs.  The first four rows use ten matched episodes;
the reward-oriented rows use the 50-episode benchmark protocol.}
\label{tab:linear}
\small
\begin{tabular}{lrrr}
\toprule
Controller & episodes & reward & late field RMSE \\
\midrule
target replay (oracle) & 10 & $-18.192\pm1.367$ & 0.00390 \\
constant $2.5$ & 10 & $-10.877\pm1.410$ & 0.05441 \\
latent-$L^2$ MPPI $(1,0)$ & 10 & $-22.011\pm1.371$ & 0.02057 \\
KE-probe MPPI $(100,0.05)$ & 10 & $-21.228\pm2.005$ & \textbf{0.01109} \\
\midrule
latent-$L^2$ MPPI $(0.05,0.05)$ & 50 & $-12.084\pm0.857$ & 0.07650 \\
KE-probe MPPI $(1,0.05)$ & 50 & $\mathbf{-10.901\pm0.911}$ & \textbf{0.06917} \\
\bottomrule
\end{tabular}
\end{table}

\newpage
The 50-episode confirmation uses seeds 0--49 and reruns the latent baseline on
the identical seeds.  KE improves reward by $1.18\pm0.20$ paired points (95\%
CI), with higher reward in 49 of 50 episodes, and lowers late field RMSE in all
50. 

For comparison with the controllers reported in PDE Control Gym, we evaluate
on its default two-dimensional Navier--Stokes tracking task
\cite{bhan2024pdecontrolgym}. The system begins from the quiescent initial
condition
\begin{equation}
    \mathbf{u}(x,y,0)=(0,0)
\end{equation}

on the unit-square domain and is simulated for $0.2\,\mathrm{s}$. The target
trajectory $\mathbf{u}^{\star}(\mathbf{x},t)$ is generated by a spatially uniform tangential boundary action applied for the full duration of the experiment:
\begin{equation}
    a^{\star}(t)=3-5t.
\end{equation}
Performance is measured by reconstructing the true trajectory. Predicted controls are applied through the native PDE solver, and the native
benchmark reward is computed as in Equation~\eqref{eq:reward}. 

We evaluate both JEPA-based planners over 50 seeded episodes, comparing their
mean episodic rewards with the 50-episode results reported in PDE Control Gym
Table~3. These values provide a benchmark-relative comparison rather than a
strictly controlled algorithmic ranking, as the published controllers update
their actions at every solver step, whereas our controller holds each macro
action for two solver steps.

\begin{table}[htbp]
\centering
\caption{Native-reward comparison on the default linear target. Baseline values are taken from PDE Control Gym Table~3
\cite{bhan2024pdecontrolgym}; for JEPA planners, we present the reward-matched mean over 50 seeded
episodes.}
\label{tab:rl}
\small
\begin{tabular}{lrr}
\toprule
Method & reward & episodes\\
\midrule
PPO & $\mathbf{-5.370}$& 50\\
Model-based optimization & $-7.931$ & 50\\
\textbf{Masked-JEPA-MPC, KE probe (ours)} & $\mathbf{-10.901}$ & 50\\
Masked-JEPA-MPC, latent-$L^2$ (ours) & $-12.084$ & 50\\
SAC & $-17.829$ & 50\\
\bottomrule
\end{tabular}
\end{table}

The observable-aligned controller improves native reward by $6.928$ relative
to the published SAC anchor and lies $2.970$ behind the published model-based
optimizer.  Published agents act every solver step whereas our action is held
for two steps, so this ranking is not final. Nevertheless, we show that a world
model trained without any reward supervision or field reconstruction can achieve competitive performance on an arbitrary control task.

\subsubsection{Three distinct aperiodic targets}\label{sec:pulse}

We next replace the benchmark reference with three deterministic, aperiodic
trajectories generated by Gaussian-pulse boundary actions. Each control signal is composed of a series of Gaussians of the form:
\begin{equation}
G(t;\mu,\sigma)
=
\exp\left[-\frac{1}{2}
\left(\frac{t-\mu}{\sigma}\right)^2\right].
\end{equation}
The specific generating functions for the aperiodic signals are included in Appendix~\ref{app:targets}. 
These schedules respectively form an up--down--up triplet, a staggered
multiscale sequence, and an alternating asymmetric sequence. All remain within
the actuator range $[0,4]$, and their pairwise action RMSEs range from $0.844$
to $0.993$. The corresponding target fields
$\mathbf{u}_k^\star(\mathbf{x},t)$ are generated by rolling each action schedule
$a_k(t)$ through the Navier--Stokes simulator.

\begin{figure}[htbp]
  \centering
  \includegraphics[width=0.94\textwidth]{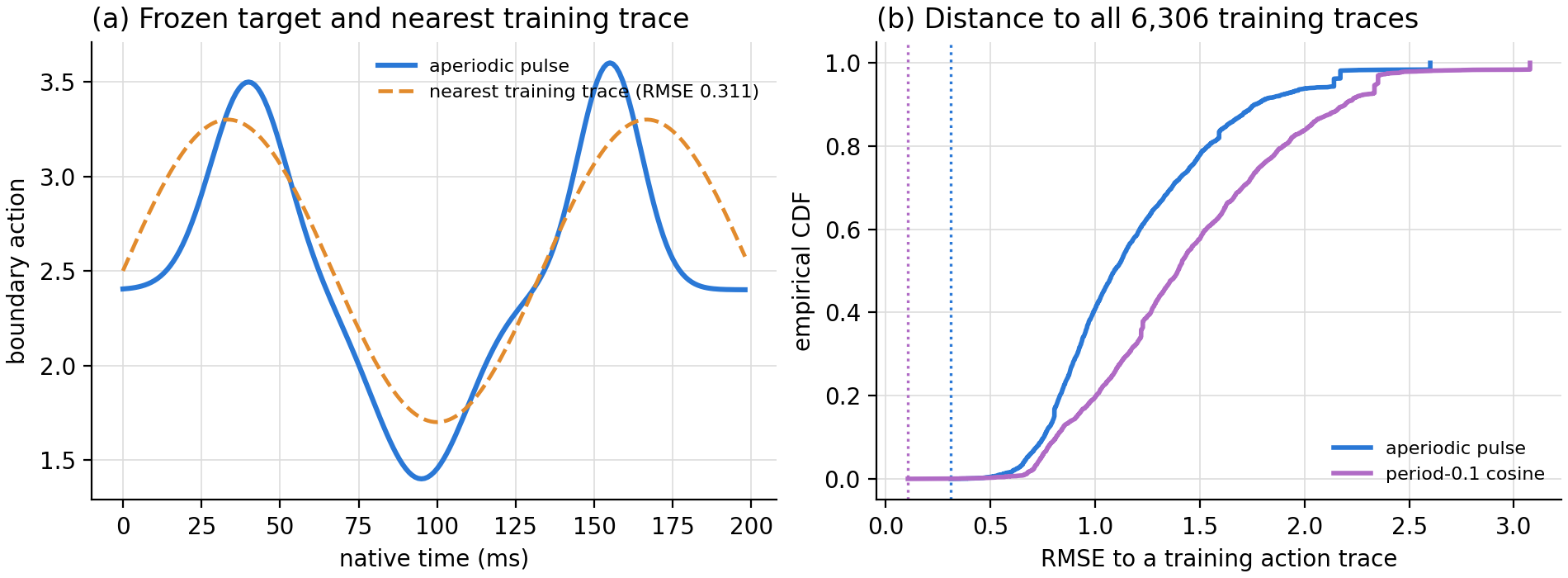}
  \caption{Action-space out-of-distribution audit for the aperiodic triplet.
  (a): the frozen triplet schedule and its nearest training action trace
  (RMSE $0.311$).  (b): empirical CDF of RMSE distances from the pulse and
  from a period-$0.1$ cosine comparator to all 6,306 training traces;
  dotted lines mark the nearest-trace distances.}
  \label{fig:pulse-ood}
\end{figure}

To quantify separation from the finite training corpus, we compare the
simplest schedule, $a_A(t)$, against all $6{,}306$ training sequences. Its
nearest-neighbor action RMSE is $0.311$, showing that it is neither present in
nor closely duplicated by the training corpus under this metric
(Figure~\ref{fig:pulse-ood}).

\begin{figure}[htbp]
  \centering
  \includegraphics[width=0.99\textwidth]{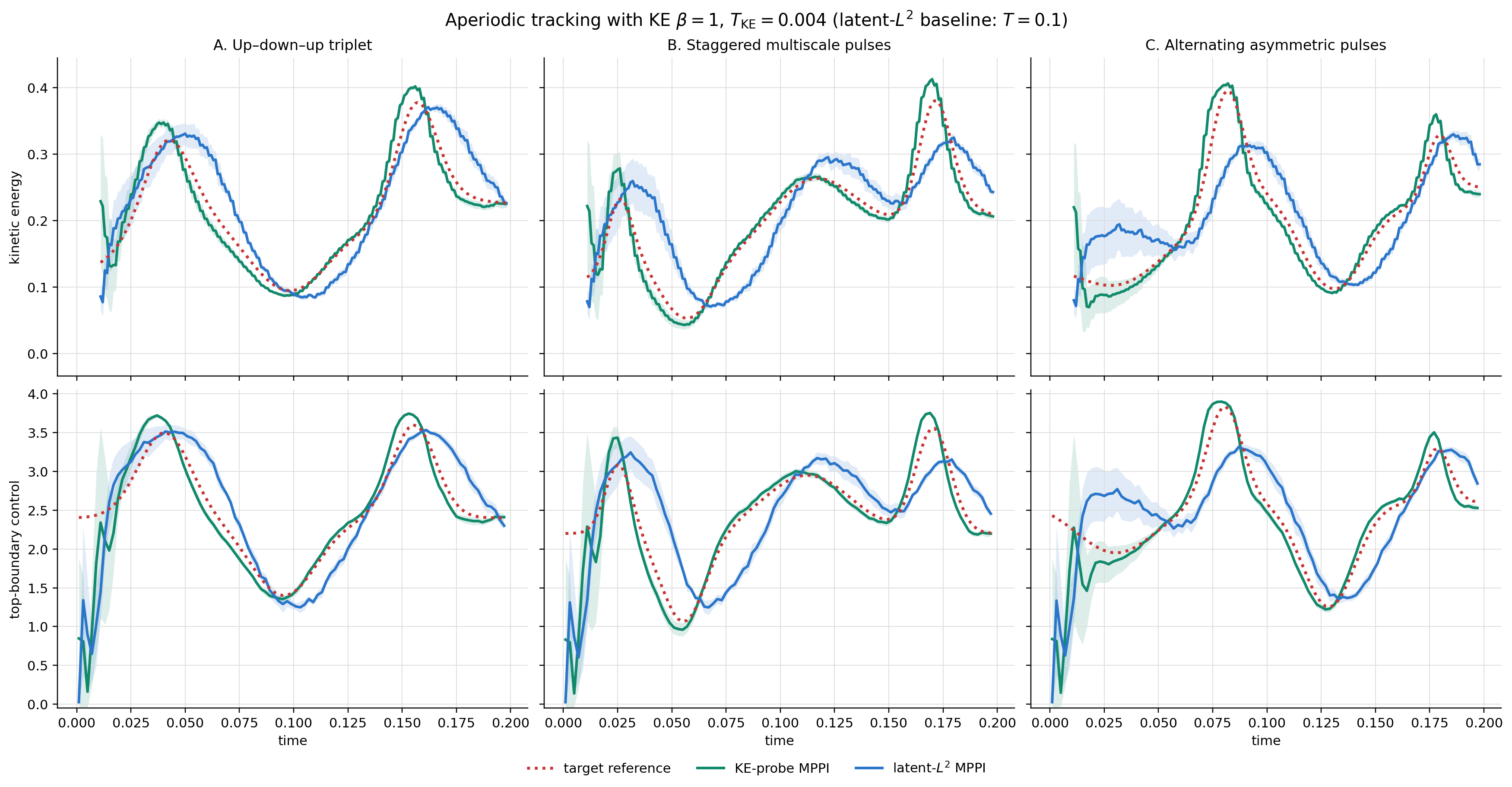}
  \caption{Clean tracking of three aperiodic targets over ten matched initial
  conditions per signal (mean with 95\% confidence bands).  Columns show the
  triplet, staggered multiscale, and alternating asymmetric targets.  The top
  row compares target and achieved KE after the first ten native steps; the
  bottom row compares the target-generating and executed boundary controls.
  Target controls are shown only for evaluation and are not supplied to MPPI.
  KE-probe planning follows their turning points, whereas latent-$L^2$
  planning exhibits phase lag.}
  \label{fig:pulse}
\end{figure}

As Figure~\ref{fig:pulse} and Table~\ref{tab:pulse} show, KE-probe planning drastically improves upon, and corrects nearly all of the phase lag observed when using latent-$L^2$ planning. Across the 30 matched episodes, it lowers late field RMSE by $53.0\%$
($0.02204$ versus $0.04687$), demonstrating a measurable improvement across all 30 episodes. The individual late-RMSE reductions are $58.2\%$,
$49.9\%$, and $50.6\%$ for the triplet, staggered, and alternating signals, shown in Figure~\ref{fig:pulse}.

\begin{table}[htbp]
\centering
\caption{Performance on three aperiodic target trajectories, each evaluated
from ten seeded initial conditions with full-state observations. Pooled
statistics weight all 30 episodes equally. Reward is reported in the native
PDE Control Gym units; late-field RMSE is the mean instantaneous full-field
RMSE over the final quarter of each episode; final-field RMSE is the
full-field error at the terminal timestep; and KE RMSE measures the discrepancy
between the controlled and target kinetic-energy trajectories.}
\label{tab:pulse}
\scriptsize
\begin{tabular}{llrrrr}
\toprule
Signal & Controller & reward & late field RMSE & final field RMSE &
late KE RMSE\\
\midrule
Triplet & latent-$L^2$ & $-16.787$ & 0.04872 & 0.02476 & 0.05097  \\
 & KE probe & $-16.044$ & \textbf{0.02037} & \textbf{0.01110} &
\textbf{0.02083}  \\
Staggered & latent-$L^2$ & $-14.502$ & 0.04963 & 0.04048 & 0.04730  \\
 & KE probe & $-14.721$ & \textbf{0.02486} & \textbf{0.00980} &
\textbf{0.02646}  \\
Alternating & latent-$L^2$ & $-13.963$ & 0.04224 & 0.03812 & 0.03845  \\
 & KE probe & $-14.970$ & \textbf{0.02088} & \textbf{0.01330} &
\textbf{0.02006}  \\
\midrule
Pooled & latent-$L^2$ & $-15.084$ & 0.04687 & 0.03445 & 0.04557  \\
 & KE probe & $-15.245$ & \textbf{0.02204} & \textbf{0.01140} &
\textbf{0.02245}  \\
\bottomrule
\end{tabular}
\end{table}

\begin{figure}[htbp]
    \centering
    \includegraphics[width=0.85\linewidth]{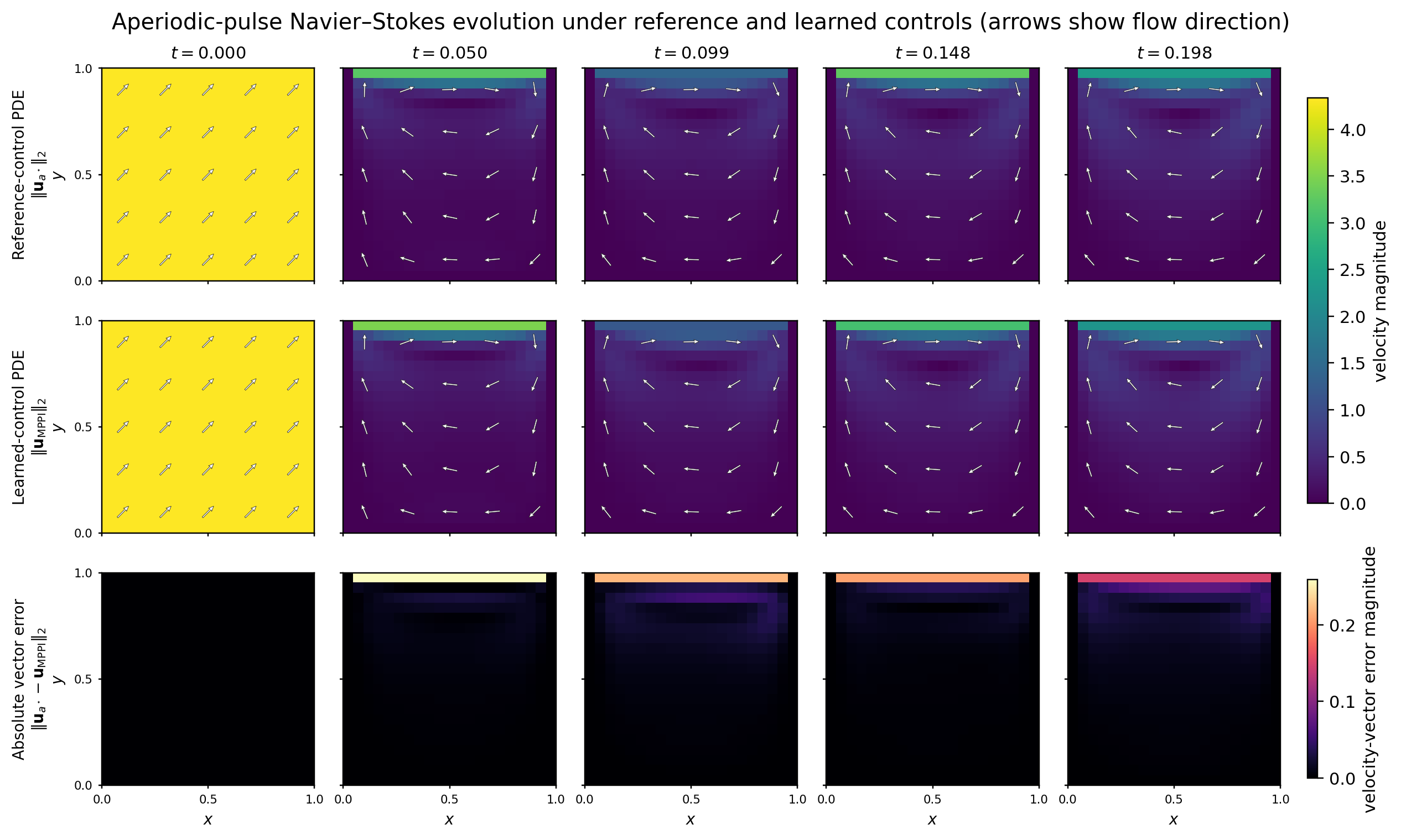}
    \caption{Aperiodic pulse reconstruction for case A: Up-down-up triplet, where both the learned and true trajectory can be shown to bear significant similarity, reflecting the accuracy with which the generating control was able to be recovered.}
    \label{fig:reconstruction}
\end{figure}

The two MPPI costs have essentially equal pooled native reward
($-15.245$ versus $-15.084$).  KE planning uses $10.6\%$ greater action total
variation, consistent with the sharper control reversals visible in the bottom
row of Figure~\ref{fig:pulse}.  The gain is therefore specifically a
physical-tracking result, not a reward claim. Purely for the sake of visualizing the field agreement, we reconstructed the first aperiodic signal and apply both the generating (oracle) and learned (JEPA) control, the results are shown in Figure~\ref{fig:reconstruction}.

\newpage

\subsection{Fixed kinetic-energy regulation}\label{sec:ke}

The tracking experiments above use the probe of Section~\ref{sec:probe} with a
time-varying reference $k(U^\star_{t+h})$.  Here we instead hold the requested
kinetic energy $k^\star$ constant, so that Equation~\eqref{eq:observable-cost}
becomes scalar setpoint regulation.  Fitted by ridge regression on frozen
predictor-rollout latents, the probe attains $R^2=0.9891$ on held-out
trajectories; the results below test whether that accuracy survives being
placed inside the planning loop.

\begin{figure}[htbp]
  \centering
  \includegraphics[width=0.84\textwidth]{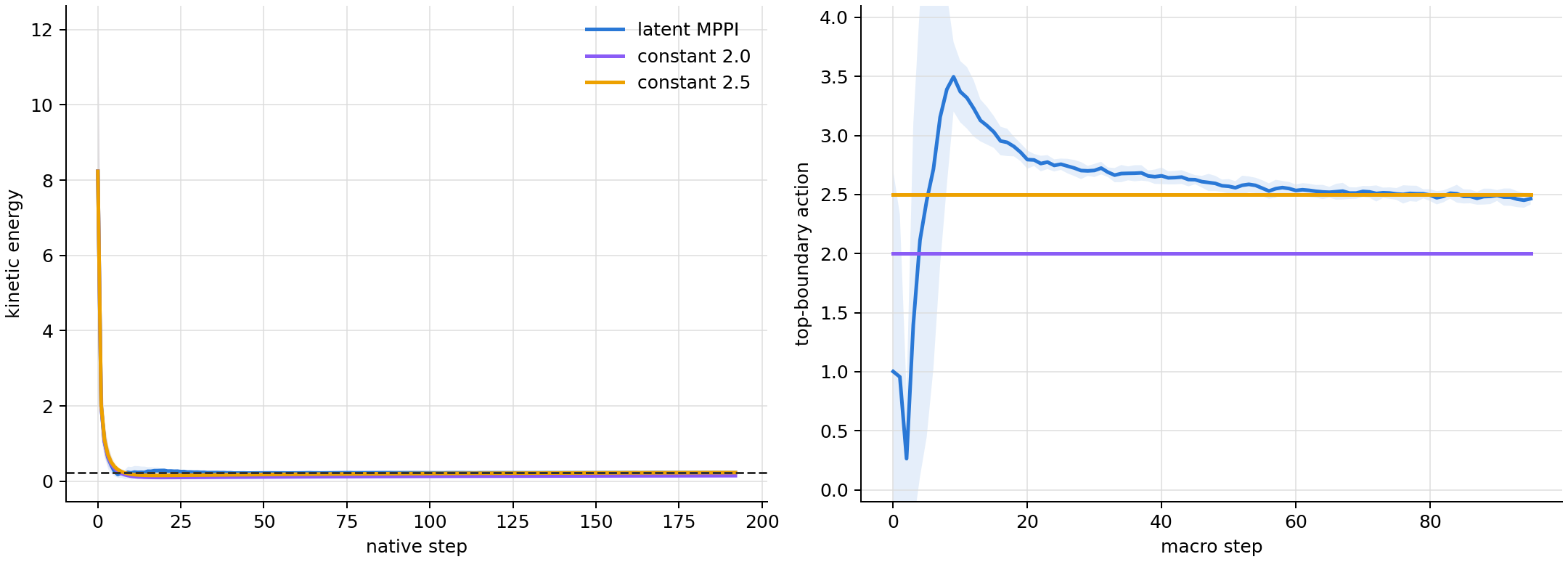}
  \caption{Derived-quantity control with a rollout-latent kinetic-energy probe.
  From ten random initial conditions, MPPI drives kinetic energy toward the
  scalar target $k^\star=0.232$ (left) and converges to a mean action of
  $2.49$, close to the true equilibrium action $2.5$ (right).}
  \label{fig:ke}
\end{figure}

Over the last 50 native steps the controller has $2.66\%$ mean pointwise
relative error, against $37.7\%$ for the constant action-reference at $2.0$,
and discovers a mean steady action of $2.491\pm0.044$ from the scalar energy
target alone --- closely recovering the true equilibrium action of $2.5$.
Running the same stabilization task through the latent-$L^2$ interface, with a
single encoded target field in place of the scalar, gives a steady field RMSE
of $0.022$.  Both interfaces therefore regulate the system successfully when
the field is fully observed.
\subsection{Limitation: probe fragility under corrupted observations}
\label{sec:robustness}

The observable-aligned tracking gains above are full-state observation results.
This qualification is important: the kinetic energy probe is demonstrably unresilient to corrupted measurements as implemented. We diagnose that failure using the fixed-KE task. Gaussian
noise is drawn per step with standard deviation equal to $5$, $10$, or $15\%$
of each velocity channel's recent RMS; fixed random-pixel masks remove $5$,
$15$, or $25\%$ of pixels. Corruptions affect only the controller's
measurement, while the simulator and target remain clean.  Each cell uses the
same ten seeds.

The fixed-field latent goal provides context rather than a headline robustness
claim: its steady RMSE rises from $0.022$ clean to $0.082$ at $15\%$ noise.
The probe-defined KE interface is much more brittle.  Its clean
$2.7\%$ relative error jumps to $46$--$64\%$ under noise and $60$--$85\%$
under missing pixels, with a large steady-action bias.  The ridge probe is
fit on clean rollout latents, and corrupted measurements evidently shift the
encoded state off the manifold on which the probe is calibrated.  Robustness
is therefore task dependent, while the frozen dynamics can accept masked or incomplete
measurements, the learned downstream readout must also be calibrated. As such, we do not claim robustness to partial or corrupted observations for the KE controller. A short
corruption diagnostic is deferred to Appendix~\ref{sec:corruption-appendix}.

\section{Discussion}\label{sec:discussion}

The experiments demonstrate the potential for JEPA world models to serve as goal-agnostic, flexible controllers for PDE problems. We show that the learned latent representation need not itself be the task metric, as JEPA can be relied upon primarily to supply action-conditioned state evolution. For our use case, where kinetic energy determines state exactly, a small readout can lead to a landscape where the control objective
is better defined.  This separation of task and latent representation preserves the claim of goal independence while avoiding the stronger and unnecessary assumption that raw latent Euclidean distance is metric preserving.

Changing the cost from minimizing $L^2$ distance to KE moves the same predictor from $-12.08$ to $-10.90$ mean reward in the 50-episode comparison. Across each of the aperiodic signals, the KE objective substantially improves control discovery, removing a phase lag present under latent-$L^2$ and halves late field error in
all 30 matched episodes. These gains indicate that the earlier bottleneck was the geometry of the cost function at least as much as the structure of the world model itself.

Across a broad class of problems, kinetic energy is not to be considered a substitute for physical field distance. We infer that our approach succeeds here because a
single scalar boundary actuator (confined to positive values only!) restricts the evaluated trajectories to a
narrow response family where the map from state to kinetic energy is actually injective.
Velocity-field RMSE is therefore retained as an independent evaluation metric,
and the result should not be extrapolated to multi-actuator or topologically
distinct flows without further observables. We do however see this as potentially positive evidence toward the claim that for well-understood systems, controlling a carefully selected observable which is known to guarantee unique continuation may prove a generalizable approach to overcoming the shortcomings of $L^2$ distance.

Our second proposed limitation is the requirement for full state observation. Since our probe is fitted on clean
rollout latents it fails when noise or missing pixels move the current-state
encoding even slightly away from the true embedding. Our observable-aligned planning in its current form thus trades an uncalibrated latent metric for an explicit readout that must itself be made
robust. Training readouts on corrupted latents, adding uncertainty quantification or other robustness measures offer avenues for natural extension.

Every experiment still uses the same five-step latent rollout.  No task-wise
horizon tuning is needed for the reported results, although longer horizons
could alter individual operating points.  The central comparison is unchanged:
the encoder, predictor, and MPPI optimizer are held fixed while only the
deployment-time cost interface changes.

\section{Conclusions}

Goal-agnostic joint-embedding control separates learning action-conditioned dynamics from control objectives. From our observations, the strongest results are achieved when the control objective can be described deterministically in terms of an explicit physical observable rather than raw latent distance. We find that a rank-one KE probe improves performance for the 
linear-reference reward, field RMSE, reduces time-dependent tracking error by
$53\%$ across three aperiodic targets relative to latent-$L^2$ alone. The same checkpoint and short-horizon planner retain fixed-field stabilization as a separate goal interface.

While KE results are not able to be generalized across settings, and
the clean-calibrated probe is brittle under corrupted observations, we believe a properly prescribed observable, if capable of being faithfully reconstructed from the latent space, could transfer the benefits found in our work. Our experiments show that reconstruction-free latent dynamics are nonetheless a powerful substrate for control, while a carefully selected physics probe can supply a better control objective, even if state control is the primary goal. The practical lesson is not that
every PDE should be controlled by energy, but that JEPA world models are capable of achieving competitive control results for PDEs in a goal-agnostic setting, and furthermore that if available, control through a calibrated observable can improve state-space tracking.

\clearpage
\appendix

\section{Hyperparameter Summary}

\begin{table}[htbp]
\centering
\caption{Principal model and planner settings.}
\small
\begin{tabular}{lll}
\toprule
Component & setting & value\\
\midrule
Encoder & latent dimension / patch / depth / heads & $256 / 3 / 4 / 4$\\
Predictor & history / training rollout / stride & $4 / 12 / 2$ native steps\\
Predictor & hidden layers / delta prediction & $[512,512]$ / yes\\
Masking & full-observation probability & $0.15$\\
Masking & masked-ratio range / pattern weights & $[0.3,0.9]$ / $[0.5,0.3,0.2]$\\
Training & steps / batch / learning rate & $50{,}000 / 128 / 3\times10^{-4}$\\
Training & continuation steps / peak learning rate  & $50{,}000 / 6\times10^{-5}$\\
VICReg & variance / covariance / $\gamma$ & $1 / 1 / 0.4$\\
Straightening & weight & $0.1$\\
Action decoder & weight & $10$\\
MPPI & samples / iterations / noise & $512 / 3 / 0.6$\\
MPPI & horizon / tail / base temperature & $5 / 5 / 0.1$\\
MPPI & latent $\alpha/\lambda_a$ (tracking / reward) & $1/0$ / $0.05/0.05$\\
MPPI & KE $(\beta,\lambda_a,\lambda_T)$ (aperiodic / reward) &
$(1,0,0.004)$ / $(1,0.05,0.1)$\\
MPPI & KE linear scale screen & $\beta\in[0.25,200]$\\
\bottomrule
\end{tabular}
\end{table}

\clearpage
\section{Corrupted-Observation Diagnostic}\label{sec:corruption-appendix}

\begin{figure}[htbp]
  \centering
  \includegraphics[width=0.9\textwidth]{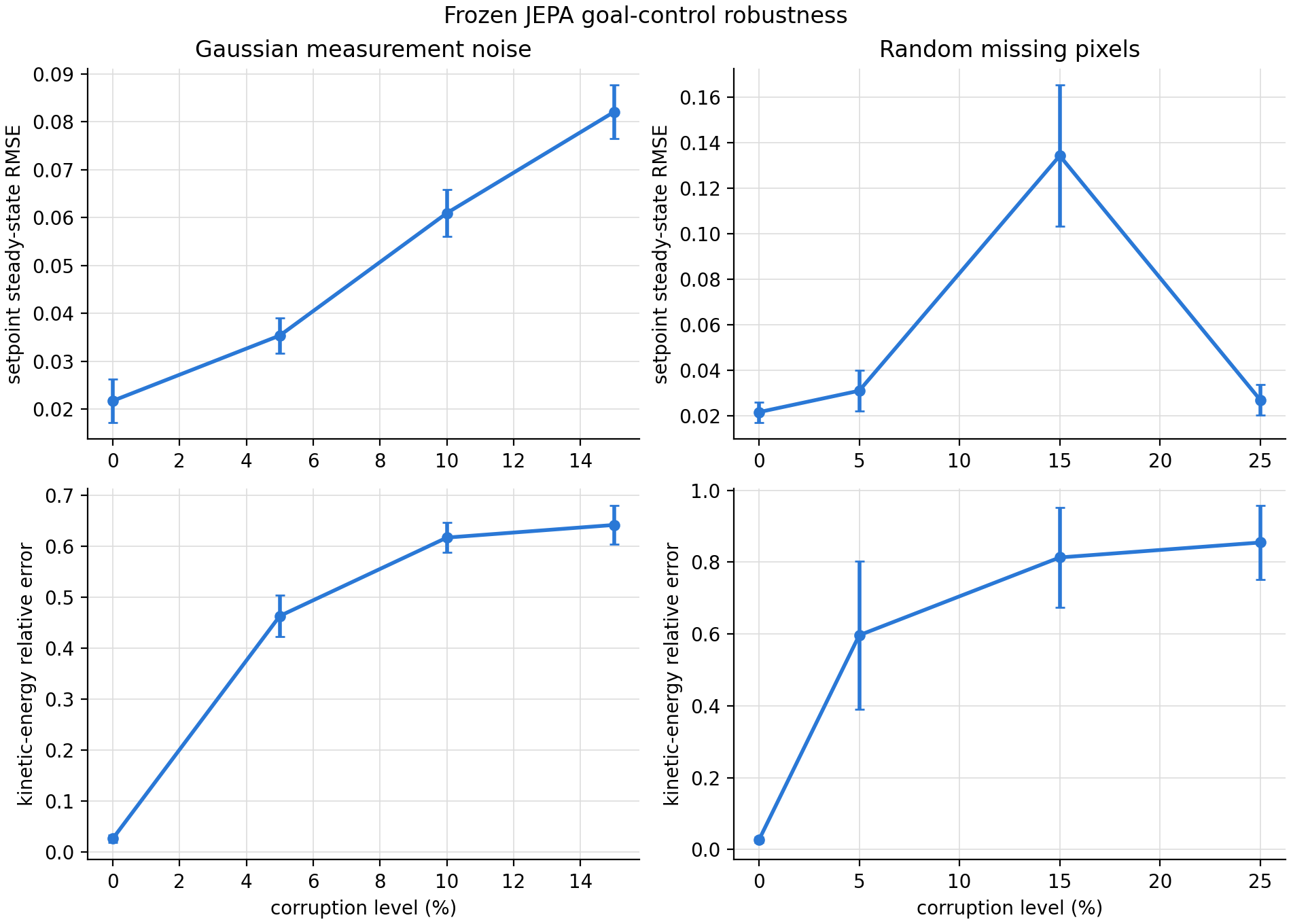}
  \caption{Secondary corruption diagnostic over ten matched episodes.
  Fixed-field stabilization remains usable across most conditions, whereas
  the clean-calibrated kinetic energy probe fails under both measurement noise and missing
  pixels (mean $\pm$ 95\% CI).}
  \label{fig:robust-goal}
\end{figure}

\section{Aperiodic target trajectories}
\label{app:targets}

The three aperiodic references of Sec.~\ref{sec:pulse} are deterministic
superpositions of Gaussian pulses applied at the boundary. Writing the
unit-amplitude pulse as
\begin{equation}
G(t;\mu,\sigma)
=\exp\!\left[-\frac{1}{2}\left(\frac{t-\mu}{\sigma}\right)^{2}\right],
\qquad \mu,\sigma>0,
\label{eq:gaussian-pulse}
\end{equation}
each generating signal takes the form
\begin{equation}
a_X(t)=a_0^{(X)}+\sum_{k=1}^{K_X} c_k^{(X)}\,G\!\left(t;\mu_k^{(X)},\sigma_k^{(X)}\right),
\qquad X\in\{A,B,C\},
\label{eq:target-general}
\end{equation}
with $K_A=3$ and $K_B=K_C=4$. Written out, the three schedules are
\begin{align}
a_A(t)&=2.4
 +1.10\,G(t;0.040,0.012)
 -1.00\,G(t;0.095,0.015) \nonumber\\
&\qquad
 +1.20\,G(t;0.155,0.010),
\label{eq:target-a}\\[4pt]
a_B(t)&=2.2
 +0.90\,G(t;0.025,0.006)
 -1.15\,G(t;0.055,0.010) \nonumber\\
&\qquad
 +0.75\,G(t;0.112,0.020)
 +1.35\,G(t;0.170,0.008),
\label{eq:target-b}\\[4pt]
a_C(t)&=2.6
 -0.65\,G(t;0.030,0.018)
 +1.25\,G(t;0.080,0.008) \nonumber\\
&\qquad
 -1.35\,G(t;0.128,0.012)
 +0.70\,G(t;0.178,0.006).
\label{eq:target-c}
\end{align}
The same parameters are collected in Table~\ref{tab:target-params} for
convenience.

\begin{table}[htbp]
\centering
\caption{Pulse parameters defining the three aperiodic targets in
Eqs.~\eqref{eq:target-a}--\eqref{eq:target-c}. Amplitudes $c_k$ are signed;
$\mu_k$ and $\sigma_k$ are in units of $t$.}
\label{tab:target-params}
\begin{tabular}{llrrr}
\toprule
Target & Baseline $a_0$ & $c_k$ & $\mu_k$ & $\sigma_k$ \\
\midrule
$A$ & $2.4$ & $+1.10$ & $0.040$ & $0.012$ \\
    &       & $-1.00$ & $0.095$ & $0.015$ \\
    &       & $+1.20$ & $0.155$ & $0.010$ \\
\midrule
$B$ & $2.2$ & $+0.90$ & $0.025$ & $0.006$ \\
    &       & $-1.15$ & $0.055$ & $0.010$ \\
    &       & $+0.75$ & $0.112$ & $0.020$ \\
    &       & $+1.35$ & $0.170$ & $0.008$ \\
\midrule
$C$ & $2.6$ & $-0.65$ & $0.030$ & $0.018$ \\
    &       & $+1.25$ & $0.080$ & $0.008$ \\
    &       & $-1.35$ & $0.128$ & $0.012$ \\
    &       & $+0.70$ & $0.178$ & $0.006$ \\
\bottomrule
\end{tabular}
\end{table}

\clearpage

\begingroup
\small
\bibliographystyle{plainnat}
\bibliography{references}

@article{bhan2024pdecontrolgym,
  title        = {{PDE Control Gym}: A Benchmark for Data-Driven Boundary Control of Partial Differential Equations},
  author       = {Bhan, Luke and Bian, Yuexin and Krstic, Miroslav and Shi, Yuanyuan},
  journal      = {arXiv preprint arXiv:2405.11401},
  year         = {2024},
  url          = {https://arxiv.org/abs/2405.11401}
}

@inproceedings{assran2023ijepa,
  title        = {Self-Supervised Learning from Images with a Joint-Embedding Predictive Architecture},
  author       = {Assran, Mahmoud and Duval, Quentin and Misra, Ishan and Bojanowski, Piotr and Vincent, Pascal and Rabbat, Michael and LeCun, Yann and Ballas, Nicolas},
  booktitle    = {Proceedings of the IEEE/CVF Conference on Computer Vision and Pattern Recognition},
  year         = {2023},
  url          = {https://arxiv.org/abs/2301.08243}
}

@inproceedings{bardes2022vicreg,
  title        = {{VICReg}: Variance-Invariance-Covariance Regularization for Self-Supervised Learning},
  author       = {Bardes, Adrien and Ponce, Jean and LeCun, Yann},
  booktitle    = {International Conference on Learning Representations},
  year         = {2022},
  url          = {https://arxiv.org/abs/2105.04906}
}

@inproceedings{williams2017mppi,
  title        = {Information Theoretic {MPC} for Model-Based Reinforcement Learning},
  author       = {Williams, Grady and Wagener, Nolan and Goldfain, Brian and Drews, Paul and Rehg, James M. and Boots, Byron and Theodorou, Evangelos A.},
  booktitle    = {2017 IEEE International Conference on Robotics and Automation},
  pages        = {1714--1721},
  year         = {2017},
  doi          = {10.1109/ICRA.2017.7989202},
  url          = {https://ieeexplore.ieee.org/document/7989202}
}

@article{zhou2024dinowm,
  title        = {{DINO-WM}: World Models on Pre-trained Visual Features Enable Zero-Shot Planning},
  author       = {Zhou, Gaoyue and Pan, Hengkai and LeCun, Yann and Pinto, Lerrel},
  journal      = {arXiv preprint arXiv:2411.04983},
  year         = {2024},
  url          = {https://arxiv.org/abs/2411.04983}
}

@article{wu2022lepde,
  title        = {Learning to Accelerate Partial Differential Equations via Latent Global Evolution},
  author       = {Wu, Tailin and Maruyama, Takashi and Leskovec, Jure},
  journal      = {arXiv preprint arXiv:2206.07681},
  year         = {2022},
  url          = {https://arxiv.org/abs/2206.07681}
}

@article{zhang2026deltajepa,
  title        = {{Delta-JEPA}: Learning Action-Sensitive World Models via Latent Difference Decoding},
  author       = {Zhang, Zhenghao and Wang, Yuanxiang and Guan, Zhenyu and Yang, Yujia and Shi, Bingkang and Zong, Tianyu and Yi, Hongzhu and Chao, Guoqing and Chen, Xingchen and Yang, Tiankun and Bao, Chenxi and Yu, Tao and Zhou, Jingjing and Xu, Jungang},
  journal      = {arXiv preprint arXiv:2606.31232},
  year         = {2026},
  url          = {https://arxiv.org/abs/2606.31232}
}

@article{maes2026leworldmodel,
  title        = {{LeWorldModel}: Stable End-to-End Joint-Embedding Predictive Architecture from Pixels},
  author       = {Maes, Lucas and Le Lidec, Quentin and Scieur, Damien and LeCun, Yann and Balestriero, Randall},
  journal      = {arXiv preprint arXiv:2603.19312},
  year         = {2026},
  url          = {https://arxiv.org/abs/2603.19312}
}

@misc{wang2026temporalstraighteninglatentplanning,
      title={Temporal Straightening for Latent Planning}, 
      author={Ying Wang and Oumayma Bounou and Gaoyue Zhou and Randall Balestriero and Tim G. J. Rudner and Yann LeCun and Mengye Ren},
      year={2026},
      eprint={2603.12231},
      archivePrefix={arXiv},
      primaryClass={cs.LG},
      url={https://arxiv.org/abs/2603.12231}, 
}

@misc{rao2026skyjepalearninglonghorizonworld,
      title={SkyJEPA: Learning Long-Horizon World Models for Zero-Shot Sim-to-Real Control of Quadrotors},
      author={Pratyaksh Rao and Wancong Zhang and Randall Balestriero and Yann LeCun and Giuseppe Loianno},
      year={2026},
      eprint={2606.23444},
      archivePrefix={arXiv},
      primaryClass={cs.RO},
      url={https://arxiv.org/abs/2606.23444},
}

@inproceedings{watter2015e2c,
  title        = {Embed to Control: A Locally Linear Latent Dynamics Model for Control from Raw Images},
  author       = {Watter, Manuel and Springenberg, Jost Tobias and Boedecker, Joschka and Riedmiller, Martin},
  booktitle    = {Advances in Neural Information Processing Systems},
  year         = {2015},
  url          = {https://arxiv.org/abs/1506.07365}
}

@article{ha2018worldmodels,
  title        = {World Models},
  author       = {Ha, David and Schmidhuber, J{\"u}rgen},
  journal      = {arXiv preprint arXiv:1803.10122},
  year         = {2018},
  url          = {https://arxiv.org/abs/1803.10122}
}

@inproceedings{hafner2019planet,
  title        = {Learning Latent Dynamics for Planning from Pixels},
  author       = {Hafner, Danijar and Lillicrap, Timothy and Fischer, Ian and Villegas, Ruben and Ha, David and Lee, Honglak and Davidson, James},
  booktitle    = {International Conference on Machine Learning},
  year         = {2019},
  url          = {https://arxiv.org/abs/1811.04551}
}

@inproceedings{hafner2020dreamer,
  title        = {Dream to Control: Learning Behaviors by Latent Imagination},
  author       = {Hafner, Danijar and Lillicrap, Timothy and Ba, Jimmy and Norouzi, Mohammad},
  booktitle    = {International Conference on Learning Representations},
  year         = {2020},
  url          = {https://arxiv.org/abs/1912.01603}
}

@inproceedings{chua2018pets,
  title        = {Deep Reinforcement Learning in a Handful of Trials Using Probabilistic Dynamics Models},
  author       = {Chua, Kurtland and Calandra, Roberto and McAllister, Rowan and Levine, Sergey},
  booktitle    = {Advances in Neural Information Processing Systems},
  year         = {2018},
  url          = {https://arxiv.org/abs/1805.12114}
}

@inproceedings{hansen2022tdmpc,
  title        = {Temporal Difference Learning for Model Predictive Control},
  author       = {Hansen, Nicklas and Wang, Xiaolong and Su, Hao},
  booktitle    = {International Conference on Machine Learning},
  year         = {2022},
  url          = {https://arxiv.org/abs/2203.04955}
}

@inproceedings{hansen2024tdmpc2,
  title        = {{TD-MPC2}: Scalable, Robust World Models for Continuous Control},
  author       = {Hansen, Nicklas and Su, Hao and Wang, Xiaolong},
  booktitle    = {International Conference on Learning Representations},
  year         = {2024},
  url          = {https://arxiv.org/abs/2310.16828}
}

@article{schrittwieser2020muzero,
  title        = {Mastering {Atari}, {Go}, Chess and Shogi by Planning with a Learned Model},
  author       = {Schrittwieser, Julian and Antonoglou, Ioannis and Hubert, Thomas and Simonyan, Karen and Sifre, Laurent and Schmitt, Simon and Guez, Arthur and Lockhart, Edward and Hassabis, Demis and Graepel, Thore and Lillicrap, Timothy and Silver, David},
  journal      = {Nature},
  volume       = {588},
  pages        = {604--609},
  year         = {2020},
  doi          = {10.1038/s41586-020-03051-4}
}

@misc{lecun2022path,
  title        = {A Path Towards Autonomous Machine Intelligence},
  author       = {LeCun, Yann},
  year         = {2022},
  note         = {OpenReview preprint, version 0.9.2},
  url          = {https://openreview.net/forum?id=BZ5a1r-kVsf}
}

@article{assran2025vjepa2,
  title        = {{V-JEPA 2}: Self-Supervised Video Models Enable Understanding, Prediction and Planning},
  author       = {Assran, Mahmoud and Bardes, Adrien and Fan, David and Garrido, Quentin and Howes, Russell and others},
  journal      = {arXiv preprint arXiv:2506.09985},
  year         = {2025},
  url          = {https://arxiv.org/abs/2506.09985}
}

@inproceedings{dosovitskiy2021vit,
  title        = {An Image is Worth 16x16 Words: Transformers for Image Recognition at Scale},
  author       = {Dosovitskiy, Alexey and Beyer, Lucas and Kolesnikov, Alexander and Weissenborn, Dirk and Zhai, Xiaohua and Unterthiner, Thomas and Dehghani, Mostafa and Minderer, Matthias and Heigold, Georg and Gelly, Sylvain and Uszkoreit, Jakob and Houlsby, Neil},
  booktitle    = {International Conference on Learning Representations},
  year         = {2021},
  url          = {https://arxiv.org/abs/2010.11929}
}

@inproceedings{grill2020byol,
  title        = {Bootstrap Your Own Latent: A New Approach to Self-Supervised Learning},
  author       = {Grill, Jean-Bastien and Strub, Florian and Altch{\'e}, Florent and Tallec, Corentin and Richemond, Pierre H. and Buchatskaya, Elena and Doersch, Carl and Pires, Bernardo Avila and Guo, Zhaohan Daniel and Azar, Mohammad Gheshlaghi and Piot, Bilal and Kavukcuoglu, Koray and Munos, R{\'e}mi and Valko, Michal},
  booktitle    = {Advances in Neural Information Processing Systems},
  year         = {2020},
  url          = {https://arxiv.org/abs/2006.07733}
}

@inproceedings{he2022mae,
  title        = {Masked Autoencoders Are Scalable Vision Learners},
  author       = {He, Kaiming and Chen, Xinlei and Xie, Saining and Li, Yanghao and Doll{\'a}r, Piotr and Girshick, Ross},
  booktitle    = {Proceedings of the IEEE/CVF Conference on Computer Vision and Pattern Recognition},
  year         = {2022},
  url          = {https://arxiv.org/abs/2111.06377}
}

@inproceedings{holl2020pdecontrol,
  title        = {Learning to Control {PDEs} with Differentiable Physics},
  author       = {Holl, Philipp and Koltun, Vladlen and Thuerey, Nils},
  booktitle    = {International Conference on Learning Representations},
  year         = {2020},
  url          = {https://arxiv.org/abs/2001.07457}
}

@inproceedings{hwang2022operatorcontrol,
  title        = {Solving {PDE}-Constrained Control Problems Using Operator Learning},
  author       = {Hwang, Rakhoon and Lee, Jae Yong and Shin, Jin Young and Hwang, Hyung Ju},
  booktitle    = {Proceedings of the AAAI Conference on Artificial Intelligence},
  pages        = {4504--4512},
  year         = {2022},
  url          = {https://arxiv.org/abs/2111.04941}
}

@article{mowlavi2023pinncontrol,
  title        = {Optimal Control of {PDEs} Using Physics-Informed Neural Networks},
  author       = {Mowlavi, Saviz and Nabi, Saleh},
  journal      = {Journal of Computational Physics},
  volume       = {473},
  pages        = {111731},
  year         = {2023},
  url          = {https://arxiv.org/abs/2111.09880}
}

@inproceedings{wei2024diffphycon,
  title        = {{DiffPhyCon}: A Generative Approach to Control Complex Physical Systems},
  author       = {Wei, Long and Hu, Peiyan and Feng, Ruiqi and Feng, Haodong and Du, Yixuan and Zhang, Tao and Wang, Rui and Wang, Yue and Ma, Zhi-Ming and Wu, Tailin},
  booktitle    = {Advances in Neural Information Processing Systems},
  year         = {2024},
  url          = {https://arxiv.org/abs/2407.06494}
}

@article{bhan2024neuralopbackstepping,
  title        = {Neural Operators for Bypassing Gain and Control Computations in {PDE} Backstepping},
  author       = {Bhan, Luke and Shi, Yuanyuan and Krstic, Miroslav},
  journal      = {IEEE Transactions on Automatic Control},
  volume       = {69},
  pages        = {5310--5325},
  year         = {2024},
  url          = {https://arxiv.org/abs/2302.14265}
}

@inproceedings{morton2018deepfluid,
  title        = {Deep Dynamical Modeling and Control of Unsteady Fluid Flows},
  author       = {Morton, Jeremy and Jameson, Antony and Kochenderfer, Mykel J. and Witherden, Freddie},
  booktitle    = {Advances in Neural Information Processing Systems},
  year         = {2018},
  url          = {https://arxiv.org/abs/1805.07472}
}

@article{lusch2018koopman,
  title        = {Deep Learning for Universal Linear Embeddings of Nonlinear Dynamics},
  author       = {Lusch, Bethany and Kutz, J. Nathan and Brunton, Steven L.},
  journal      = {Nature Communications},
  volume       = {9},
  pages        = {4950},
  year         = {2018},
  doi          = {10.1038/s41467-018-07210-0}
}

@article{korda2018koopmanmpc,
  title        = {Linear Predictors for Nonlinear Dynamical Systems: {Koopman} Operator Meets Model Predictive Control},
  author       = {Korda, Milan and Mezi{\'c}, Igor},
  journal      = {Automatica},
  volume       = {93},
  pages        = {149--160},
  year         = {2018},
  doi          = {10.1016/j.automatica.2018.03.046}
}

@article{peitz2019koopmanpde,
  title        = {Koopman Operator-Based Model Reduction for Switched-System Control of {PDEs}},
  author       = {Peitz, Sebastian and Klus, Stefan},
  journal      = {Automatica},
  volume       = {106},
  pages        = {184--191},
  year         = {2019},
  doi          = {10.1016/j.automatica.2019.05.016}
}

@article{kaiser2018sindympc,
  title        = {Sparse Identification of Nonlinear Dynamics for Model Predictive Control in the Low-Data Limit},
  author       = {Kaiser, Eurika and Kutz, J. Nathan and Brunton, Steven L.},
  journal      = {Proceedings of the Royal Society A},
  volume       = {474},
  number       = {2219},
  pages        = {20180335},
  year         = {2018},
  doi          = {10.1098/rspa.2018.0335}
}

@article{tomasetto2024latentfeedback,
  title        = {Latent Feedback Control of Distributed Systems in Multiple Scenarios Through Deep Learning-Based Reduced Order Models},
  author       = {Tomasetto, Matteo and Braghin, Francesco and Manzoni, Andrea},
  journal      = {arXiv preprint arXiv:2412.09942},
  year         = {2024},
  url          = {https://arxiv.org/abs/2412.09942}
}

@article{rabault2019flowcontrol,
  title        = {Artificial Neural Networks Trained Through Deep Reinforcement Learning Discover Control Strategies for Active Flow Control},
  author       = {Rabault, Jean and Kuchta, Miroslav and Jensen, Atle and R{\'e}glade, Ulysse and Cerardi, Nicolas},
  journal      = {Journal of Fluid Mechanics},
  volume       = {865},
  pages        = {281--302},
  year         = {2019},
  doi          = {10.1017/jfm.2019.62}
}

@article{verma2018collective,
  title        = {Efficient Collective Swimming by Harnessing Vortices Through Deep Reinforcement Learning},
  author       = {Verma, Siddhartha and Novati, Guido and Koumoutsakos, Petros},
  journal      = {Proceedings of the National Academy of Sciences},
  volume       = {115},
  number       = {23},
  pages        = {5849--5854},
  year         = {2018},
  doi          = {10.1073/pnas.1800923115}
}

@article{bucci2019chaotic,
  title        = {Control of Chaotic Systems by Deep Reinforcement Learning},
  author       = {Bucci, Michele Alessandro and Semeraro, Onofrio and Allauzen, Alexandre and Wisniewski, Guillaume and Cordier, Laurent and Mathelin, Lionel},
  journal      = {Proceedings of the Royal Society A},
  volume       = {475},
  number       = {2231},
  pages        = {20190351},
  year         = {2019},
  doi          = {10.1098/rspa.2019.0351}
}

@inproceedings{farahmand2017pdecontrol,
  title        = {Deep Reinforcement Learning for Partial Differential Equation Control},
  author       = {Farahmand, Amir-massoud and Nabi, Saleh and Nikovski, Daniel N.},
  booktitle    = {2017 American Control Conference},
  pages        = {3120--3127},
  year         = {2017},
  doi          = {10.23919/ACC.2017.7963427}
}

@article{brunton2015closedloop,
  title        = {Closed-Loop Turbulence Control: Progress and Challenges},
  author       = {Brunton, Steven L. and Noack, Bernd R.},
  journal      = {Applied Mechanics Reviews},
  volume       = {67},
  number       = {5},
  pages        = {050801},
  year         = {2015},
  doi          = {10.1115/1.4031175}
}
\endgroup

\end{document}